\documentclass{article}

\PassOptionsToPackage{numbers, compress}{natbib}
\usepackage[main, final]{neurips_2026}

\usepackage[utf8]{inputenc} % allow utf-8 input
\usepackage[T1]{fontenc}    % use 8-bit T1 fonts
\usepackage{hyperref}       % hyperlinks
\usepackage{url}            % simple URL typesetting
\usepackage{booktabs}       % professional-quality tables
\usepackage{graphicx}       % resize tables and figures
\usepackage{caption}        % provides \captionof outside floats
\usepackage{multirow}       % multirow cells in tables
\usepackage{amsmath}        % advanced math typesetting
\usepackage{amsfonts}       % blackboard math symbols
\usepackage{nicefrac}       % compact symbols for 1/2, etc.
\usepackage{microtype}      % microtypography
\usepackage{xcolor}         % colors
\usepackage{colortbl}
\usepackage{enumitem}
\usepackage{bbm}
\usepackage[capitalize]{cleveref}
\crefname{section}{Sec.}{Secs.}
\crefname{table}{Tab.}{Tabs.}
\crefname{figure}{Fig.}{Figs.}

\newcommand{\venue}[1]{\textcolor{gray}{\scriptsize\,#1}}
\title{PCDFusion: Proposal-Context-Detail Bayesian Rendering for Infrared-Visible Image Fusion}

\author{%
  Rui Liu \\
  School of Computer Science and Technology \\
  Beijing Institute of Technology \\
  Beijing 100081, China \\
  \texttt{liurui25@bit.edu.cn}
  \And
  Lin Gu \\
  Research Institute of Electrical Communication \\
  Tohoku University \\
  Sendai 980-8577, Japan \\
  \texttt{lin@tohoku.ac.jp}
  \AND
  Hesong Li \\
  School of Computer Science and Technology \\
  Beijing Institute of Technology \\
  Beijing 100081, China \\
  \texttt{lihesong2@bit.edu.cn}
  \And
  Ying Fu\thanks{Corresponding author.} \\
  School of Computer Science and Technology \\
  Beijing Institute of Technology \\
  Beijing 100081, China \\
  \texttt{fuying@bit.edu.cn}
}

\begin{document}
\maketitle
\begin{abstract}
Infrared-visible image fusion (IVIF) aims to combine visible and thermal infrared observations for human perception and downstream vision tasks. 
Visible and thermal infrared observations are physically complementary but unreliable in different ways.
Existing deep IVIF methods mainly improve feature interaction and reconstruction, but often lack an explicit local rule for deciding when visible appearance should dominate and when thermal responses should contribute. 
To address this issue, we reformulate IVIF as local Bayesian posterior rendering, where fusion is controlled by posterior action weights rather than direct modality mixing. 
In this paper, we propose Proposal-Context-Detail Fusion (PCDFusion), a visible-anchored framework that implements this formulation through three rendering stages. 
PCDFusion preserves potentially useful thermal responses, updates exposure-degraded visible regions, and renders reliable band-limited infrared structures into the final luminance. 
Extensive experiments on public IVIF benchmarks show that PCDFusion improves fusion quality, target recovery, and downstream detection and segmentation performance under challenging illumination. 
The code is available at \href{https://github.com/BIT-RuiLiu/PCDFusion.git}{https://github.com/BIT-RuiLiu/PCDFusion.git}.
\end{abstract}

\section{Introduction}
\label{sec:intro}

\begin{figure*}[t]
    \centering
    \includegraphics[width=\textwidth]{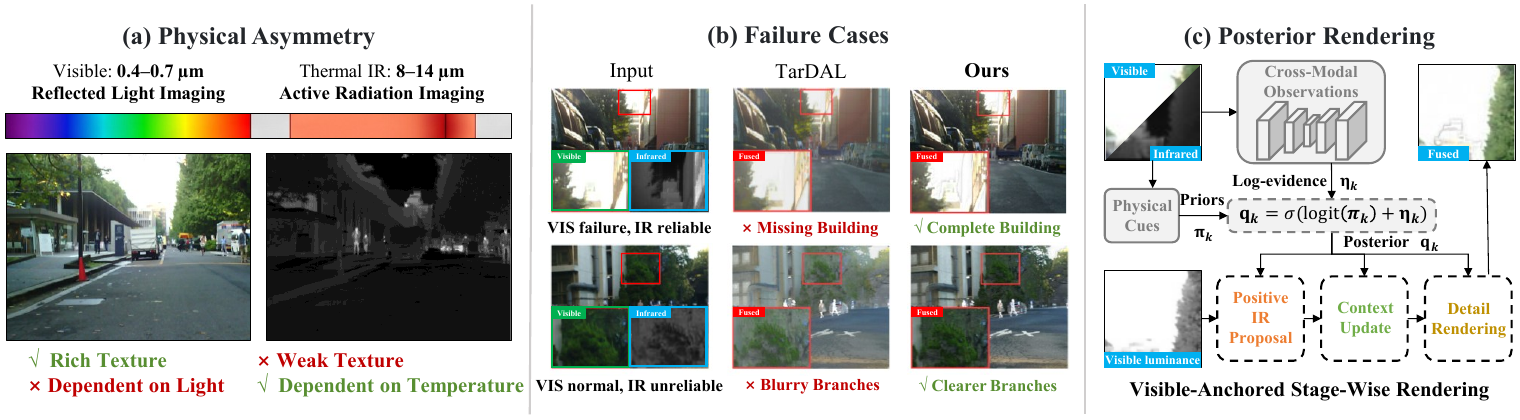}
    \caption{PCDFusion estimates posterior action weights from physical cues and cross-modal observations, recovering structures under visible failure while suppressing unreliable thermal responses.}
    \label{fig:teaser}
\end{figure*}

Infrared-visible image fusion (IVIF) aims to combine visible and thermal infrared observations into a unified image representation for human perception and downstream computer vision tasks. Visible and infrared images are physically heterogeneous sensing modalities. Visible cameras capture reflected radiance within the $0.4$--$0.7\,\mu\mathrm{m}$ spectrum \cite{schowengerdt2006remote}. 
As illustrated in \cref{fig:teaser}(a), visible images usually preserve natural appearance, rich textures, and fine structural details under proper exposure, but their quality strongly depends on lighting. In contrast, thermal infrared imaging mainly operates in the long-wave infrared band, approximately $8$--$14\,\mu\mathrm{m}$, measuring emitted radiance related to object temperature \cite{vollmer2018infrared,kuenzer2013thermal}. Thermal observations are therefore robust to darkness and illumination changes \cite{ma2019survey,zhang2023visible,ma2023technology, zhang2025unaligned}, but they often lack natural textures and reliable high-frequency structures.

Recent deep learning approaches improve IVIF through learnable representation \cite{densefuse,rfnnest,liu2024infrared}, decomposition-based modeling \cite{didfuse,cddfuse}, attention-based interaction \cite{swinfusion,iviformer,liu2026rdanet}, generative reconstruction \cite{ddfm,diffif, zhang2026supervise}, and task-driven optimization \cite{tardal,segmif,TriMSOD}. However, most methods still formulate fusion as direct feature aggregation or modality balancing, implicitly assuming symmetric contributions from visible and infrared observations despite their physical asymmetry. As shown in \cref{fig:teaser}(b), this limitation becomes evident under local reliability conflicts. When visible observations (green box) are saturated or degraded while infrared observations (blue box) retain useful structures, existing methods may still leave the failed visible region unresolved.  For example, TarDAL \cite{tardal} keeps the building region over-bright instead of recovering its structure from thermal observations. Conversely, when visible appearance remains reliable but infrared responses are unreliable for the local scene, direct fusion strategies tend to blur visible branches and introduce thermal contamination.

Motivated by this observation, we reformulate IVIF as local Bayesian posterior rendering under heterogeneous reliability. Instead of directly mixing visible and infrared features, each local rendering operation is treated as a posterior decision conditioned on physical cues and cross-modal observations. As illustrated in \cref{fig:teaser}(c), physical cues define a reliability prior, while cross-modal observations provide a learned log-evidence score (log-odds correction) for the corresponding rendering action. The prior--evidence odds update then yields a normalized posterior action weight. 
Under this formulation, thermal responses are not uniformly injected into the visible-anchored rendering. They contribute where visible observations become unreliable and are suppressed where visible appearance remains reliable.

In this paper, we propose \textbf{P}roposal-\textbf{C}ontext-\textbf{D}etail Fusion (\textbf{PCDFusion}), a visible-anchored posterior rendering framework for IVIF. PCDFusion applies the same prior--evidence--posterior update rule across three sequential rendering stages. The proposal stage preserves potentially useful thermal responses as candidates for later rendering. The context stage updates visible regions degraded by exposure failure through posterior-weighted context recovery. The detail stage selectively renders locally reliable band-limited infrared structures into the final luminance. By separating thermal proposal, context correction, and detail rendering into different posterior decisions, PCDFusion reduces thermal contamination while preserving visible appearance and structural consistency. Extensive experiments on public IVIF benchmarks demonstrate improved fusion quality, target recovery, and downstream perception under challenging illumination. Our contributions are as follows:
\begin{itemize}[leftmargin=*]
    \item We formulate IVIF as local Bayesian posterior rendering, where physical priors and learned log-evidence scores determine reliability-conditioned rendering weights.
    \item We propose PCDFusion, a visible-anchored Proposal-Context-Detail network that keeps positive thermal responses available, updates exposure-degraded visible regions, and renders band-limited thermal structures into the final luminance.
    \item Without semantic labels for fusion training, PCDFusion improves image-level fusion quality and downstream detection and segmentation performance on public benchmarks.
\end{itemize}

\section{Related Work}
\label{sec:related}

\noindent \textbf{Learning-based IVIF approaches.}
Infrared-visible image fusion has progressed from hand-crafted fusion rules to learnable multimodal representation and reconstruction. Early deep models use autoencoders, convolutional networks, dense aggregation, residual fusion, or unified unsupervised mappings to combine visible and infrared inputs \cite{densefuse,rfnnest,u2fusion, zhang2024deep, SFFN, BinRVR, Li2023FCDFusion}. Decomposition-based methods further separate shared/specific or base/detail representations to improve structure preservation and modality complementarity \cite{didfuse,cddfuse}. Recent methods enhance cross-modal interaction with Transformer-based architectures~\cite{swinfusion, li2023joint, CJE1}, cross-correlation learning~\cite{tccfusion}, adaptive cross-attention~\cite{iviformer, CJE2}, equivariant modeling~\cite{emma}, evolutionary search~\cite{dcevo}, degradation-robust denoising~\cite{denoif}, diffusion-based generation~\cite{ddfm,diffif}, and pretrained-knowledge-guided unified fusion~\cite{upfusion}. Task-driven methods additionally optimize fused images for detection, segmentation, or semantic consistency, showing the importance of downstream perception \cite{tardal,timfusion,segmif,mrfs,sage,liu2025compatibility, chen2026survey}. These methods improve feature extraction, visual reconstruction, and task utility, but they mainly focus on how visible and infrared information should be mixed. The local dominance relation between reflected-light appearance and thermal response remains mostly implicit.

\noindent \textbf{Probabilistic and prior-guided fusion.}
Probabilistic modeling and explicit priors have also been explored in image fusion. Bayesian Fusion formulates IVIF as a hierarchical Bayesian regression problem and infers the fused image with EM optimization and total-variation regularization \cite{zhao2020bayesian}. Diffusion-based fusion methods model multimodal fusion as conditional generation or denoising with task-specific fusion priors \cite{ddfm,diffif}. Model-based Bayesian fusion networks further unfold hierarchical Bayesian inference into trainable neural architectures \cite{li2025bayesianfusionnet}. In parallel, prior-guided fusion methods introduce explicit observation cues to adapt fusion under adverse imaging conditions. 
Illumination-aware methods modulate fusion according to lighting states~\cite{piafusion,iaifnet,fiafusion}, 
while darkness-aware methods specifically compensate for degraded visible observations in low-light scenes~\cite{divfusion}. These works validate probabilistic modeling and physical priors, but usually define uncertainty over the fused image or generation process, or use priors as auxiliary modulation. PCDFusion instead moves the probabilistic decision to the rendering level, estimating local posterior weights from physical priors and learned log-evidence scores.

\section{PCDFusion}
\label{sec:method}
\subsection{Formulation and Overview}
\label{sec:bayesian_rendering}
\noindent\textbf{Bayesian posterior formulation.}
Let $\mathbf{Y}_{v}$ and $\mathbf{Y}_{r}$ denote the visible luminance and thermal infrared observation, respectively. 
They are heterogeneous measurements: $\mathbf{Y}_{v}$ mainly reflects illumination-dependent appearance, while $\mathbf{Y}_{r}$ records emitted-radiance thermal responses. 
Therefore, fusion should not treat them as symmetric intensity sources. 
Instead, PCDFusion models each local rendering operation as a posterior decision conditioned on physical cues and cross-modal observations.

For the $k$-th rendering operation, we introduce a binary latent variable $z_k\in\{0,1\}$, where $z_k=1$ accepts the stage-specific rendering action and $z_k=0$ retains its base outcome. 
Let $\mathcal{P}_k(\mathbf{Y}_{v},\mathbf{Y}_{r})$ denote the physical cue used to form the prior, and let $\mathcal{O}_k$ denote the local observations used for evidence estimation. 
An ideal probabilistic model defines
\begin{gather}
\boldsymbol{\pi}_k
=
p(z_k=1\mid \mathcal{P}_k),\qquad
\boldsymbol{\ell}_k^{+}
=
p(\mathcal{O}_k\mid z_k=1,\mathcal{P}_k),\qquad
\boldsymbol{\ell}_k^{-}
=
p(\mathcal{O}_k\mid z_k=0,\mathcal{P}_k).
\end{gather}
Bayes' rule in odds form gives
\begin{gather}
\frac{\mathbf{q}_k^{\star}}{1-\mathbf{q}_k^{\star}}
=
\frac{\boldsymbol{\pi}_k}{1-\boldsymbol{\pi}_k}
\,\cdot\,
\frac{\boldsymbol{\ell}_k^{+}}{\boldsymbol{\ell}_k^{-}},
\qquad
\boldsymbol{\eta}_k^{\star}
=
\log\frac{\boldsymbol{\ell}_k^{+}}{\boldsymbol{\ell}_k^{-}},
\qquad
\mathbf{q}_k^{\star}
=
\sigma\!\left(
\operatorname{logit}(\boldsymbol{\pi}_k)+\boldsymbol{\eta}_k^{\star}
\right),
\label{eq:bayes_posterior}
\end{gather}
where $\mathbf{q}_k^{\star}=p(z_k=1\mid\mathcal{O}_k,\mathcal{P}_k)$ and $\boldsymbol{\eta}_k^{\star}$ is the exact log Bayes factor. 
In practice, PCDFusion does not fit the two class-conditional densities explicitly. 
Instead, it amortizes their relative evidence with a neural estimator:
\begin{gather}
\boldsymbol{\eta}_k
=
\smash{f_{\theta_k}(\mathcal{O}_k,\mathcal{P}_k)},
\qquad
\mathbf{B}_k=\exp(\boldsymbol{\eta}_k),
\qquad
\mathbf{q}_k
=
\sigma\!\left(
\operatorname{logit}(\boldsymbol{\pi}_k)+\boldsymbol{\eta}_k
\right)
=
\frac{\boldsymbol{\pi}_k\mathbf{B}_k}
{1-\boldsymbol{\pi}_k+\boldsymbol{\pi}_k\mathbf{B}_k}.
\label{eq:bayes_logit}
\end{gather}
Here, $\boldsymbol{\eta}_k$ is a learned log-evidence score (log-odds correction), and $\mathbf{B}_k$ is the corresponding multiplicative evidence factor.
When $\boldsymbol{\eta}_k=0$, the posterior remains at its prior value, $\mathbf{q}_k=\boldsymbol{\pi}_k$; positive and negative evidence shift the action odds above or below the physical prior.
The resulting $\mathbf{q}_k$ is a normalized posterior action weight with an explicit prior anchor and a signed evidence correction.

The posterior is used to render the expectation of two stage-specific outcomes $\mathcal{R}_{k,0}$ and $\mathcal{R}_{k,1}$:
\begin{gather*}
\widehat{\mathcal{R}}_k
=
\mathbb{E}[\mathcal{R}_k\mid\mathcal{O}_k,\mathcal{P}_k]
=
(1-\mathbf{q}_k)\mathcal{R}_{k,0}
+
\mathbf{q}_k\mathcal{R}_{k,1}.
\end{gather*}
Proposal uses $\bigl(\mathbf{Y}_{v},\,\mathbf{Y}_{v}+(1-\mathbf{Y}_{v})\odot\mathbf{G}_{p}\bigr)$, context uses $(\mathbf{Y}_{p},\mathbf{Y}_{\mathrm{ctx}})$, and detail uses $(\mathbf{Y}_{c},\mathbf{Y}_{c}+\mathbf{D}_{r})$ as the two outcomes. 
The evidence estimators are optimized end-to-end through these posterior-weighted rendering actions, without direct probabilistic supervision of the scores. Calibration is evaluated empirically on the normalized posterior action weights.
Together, the posterior update and outcome expectation form a local Bayesian rendering mechanism: $\boldsymbol{\pi}_k$ anchors the rendering decision, $\boldsymbol{\eta}_k$ adapts its log odds to local observations, and $\mathbf{q}_k$ mixes the base and action-conditioned outcomes.
The same mechanism is instantiated at the proposal, context, and detail stages.

\begin{figure*}[t]
    \centering
    \includegraphics[width=\textwidth]{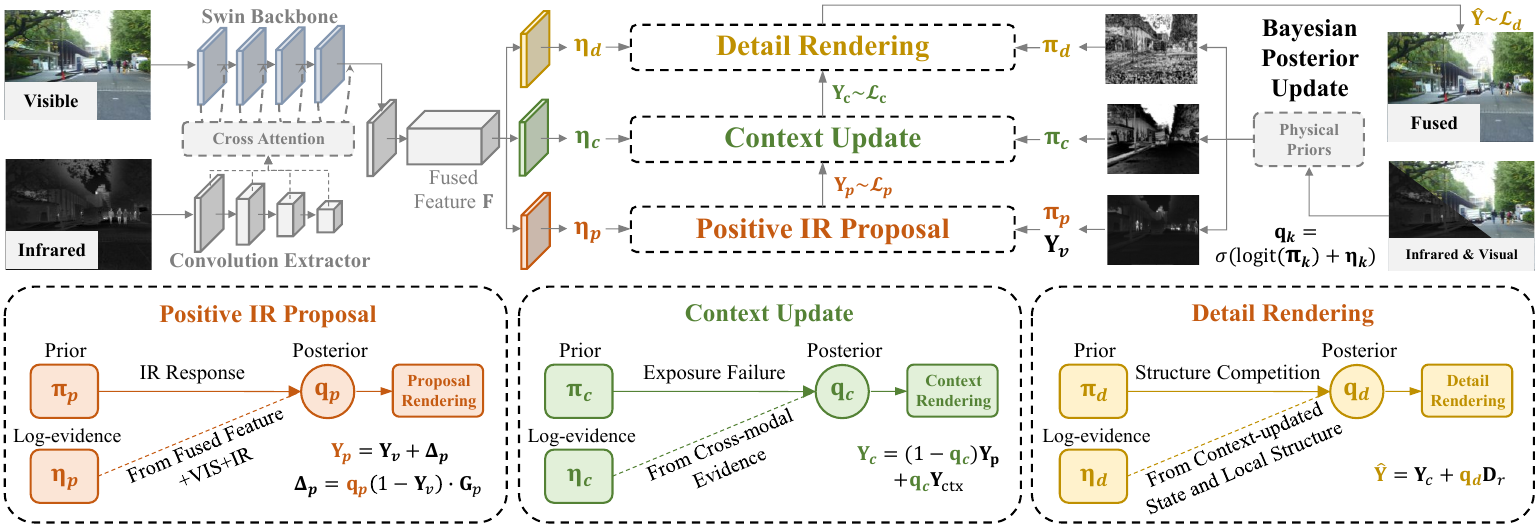}
    \caption{Overall architecture of the proposed PCDFusion.}
    \label{fig:overview}
\end{figure*}

\noindent\textbf{Architecture overview.}
As shown in \cref{fig:overview}, PCDFusion uses the visible image as the main appearance stream. 
A Swin Transformer extracts visible features, while a lightweight convolutional encoder extracts infrared features. 
The infrared features are injected into the visible stream through attention-based fusion, producing a visible-anchored representation $\mathbf{F}$. 
Based on \cref{eq:bayes_logit}, PCDFusion estimates three posterior action weights: $\mathbf{q}_{p}$ for positive thermal-response proposal, $\mathbf{q}_{c}$ for visible-context update, and $\mathbf{q}_{d}$ for infrared-detail rendering. 
The following subsections specify the corresponding priors, log-evidence estimators, and rendering equations.

\subsection{Bayesian Positive Infrared Proposal}
\label{sec:positive_proposal}

The first stage estimates whether positive thermal responses should be preserved before context and detail rendering. 
This stage is designed for high recall: potential thermal targets should remain available for later posterior decisions, but they should not be forced into the final rendering.

\noindent \textbf{Prior and posterior.}
We use the normalized infrared observation to form a thermal-response prior:
\begin{gather}
\boldsymbol{\pi}_{p}
=
p(z_{p}=1\mid \mathcal{P}_{p})
=
\sigma\!\left(
\tau_{p}
\left(
\mathrm{Norm}(\mathbf{Y}_{r})-0.5
\right)
\right),
\end{gather}
where $z_{p}=1$ denotes that positive thermal responses should be preserved, and $\tau_{p}$ controls the prior sharpness. 
Since infrared intensity alone cannot determine whether the response should be retained, PCDFusion estimates the log-evidence score from the visible-anchored feature and the two observations:
\begin{gather}
\boldsymbol{\eta}_{p}
=
f^{p}_{\theta}
(\mathbf{F},\mathbf{Y}_{v},\mathbf{Y}_{r})
,
\qquad
\mathbf{q}_{p}
=
\sigma\!\left(
\operatorname{logit}(\boldsymbol{\pi}_{p})+\boldsymbol{\eta}_{p}
\right).
\end{gather}
Here, $\boldsymbol{\eta}_{p}$ is the learned log-evidence score and $\mathbf{q}_{p}$ is the proposal posterior action weight.

\noindent \textbf{Proposal rendering.}
The proposal update should be positive and bounded by the remaining visible luminance margin. 
PCDFusion therefore predicts a conditional gain and renders the proposal as
\begin{gather}
\mathbf{G}_{p}
=
\sigma\!\left(
g^{p}_{\theta}
(\mathbf{F},\mathbf{Y}_{v},\mathbf{Y}_{r})
\right),
\qquad
\boldsymbol{\Delta}_{p}
=
\mathbf{q}_{p}
\odot
(1-\mathbf{Y}_{v})
\odot
\mathbf{G}_{p},
\qquad
\mathbf{Y}_{p}
=
\mathbf{Y}_{v}
+
\boldsymbol{\Delta}_{p}.
\end{gather}
Here, $\mathbf{G}_{p}$ is the conditional gain, $(1-\mathbf{Y}_{v})$ is the available positive luminance margin, and $\boldsymbol{\Delta}_{p}$ is the positive proposal correction.

\subsection{Bayesian Visible Context Update}
\label{sec:context_update}

After obtaining $\mathbf{Y}_{p}$, PCDFusion estimates whether the visible-anchored context should be revised. 
This stage addresses illumination-induced visible failure: under-exposure lowers the effective signal-to-noise ratio, while over-exposure causes saturation and local contrast collapse.

\noindent \textbf{Prior and posterior.}
The context prior is defined from visible luminance:
\begin{gather}
\boldsymbol{\pi}_{c}
=
p(z_{c}=1\mid \mathcal{P}_{c})
=
1-4\mathbf{Y}_{v}(1-\mathbf{Y}_{v}),
\end{gather}
where $z_{c}=1$ indicates that the visible context should be updated. 
This U-shaped prior is low under normal exposure and high near darkness or saturation. 
Since exposure failure alone does not guarantee useful correction, PCDFusion estimates the log-evidence score from cross-modal observations and structural cues:
\begin{gather}
\boldsymbol{\eta}_{c}
=
f^{c}_{\theta}
(
\mathbf{F},
\mathbf{Y}_{v},
\mathbf{Y}_{r},
\boldsymbol{\pi}_{p},
\boldsymbol{\pi}_{c},
\mathbf{S}_{v},
\mathbf{S}^{b}_{r}
)
,
\quad
\mathbf{q}_{c}
=
\sigma\!\left(
\operatorname{logit}(\boldsymbol{\pi}_{c})+\boldsymbol{\eta}_{c}
\right).
\end{gather}
Here, $\mathbf{S}_{v}$ denotes visible structural support, $\mathbf{S}^{b}_{r}$ denotes band-limited infrared structural support, $\boldsymbol{\eta}_{c}$ is the learned log-evidence score, and $\mathbf{q}_{c}$ is the context posterior action weight.

\noindent \textbf{Context rendering.}
The context action weight controls interpolation between the proposal rendering and a recovered context estimate. 
PCDFusion predicts the recovered context with a value prediction and a bounded residual refinement:
\begin{gather}
\mathbf{Y}_{\mathrm{ctx}}
=
\sigma\!\left(
\mathbf{C}_{\theta}
+
\mathbf{R}_{\theta}
\right),
\qquad
\mathbf{Y}_{c}
=
(1-\mathbf{q}_{c})\odot\mathbf{Y}_{p}
+
\mathbf{q}_{c}\odot\mathbf{Y}_{\mathrm{ctx}}.
\end{gather}
Here, $\mathbf{C}_{\theta}$ is the context value prediction, $\mathbf{R}_{\theta}$ is the bounded residual refinement, $\mathbf{Y}_{\mathrm{ctx}}$ is the recovered context, and $\mathbf{Y}_{c}$ is the context-updated rendering.

\subsection{Bayesian Infrared Detail Rendering}
\label{sec:detail_rendering}

After context update, PCDFusion estimates whether infrared detail should contribute to the final luminance. 
Raw infrared intensity is insufficient for this decision, because strong infrared responses may also come from smooth thermal background, emissivity variation, thermal reflection, or sensor artifacts. 
We therefore use local structure as the physical cue, so that infrared detail is favored only when it provides stronger local structural support than the visible anchor.

\noindent \textbf{Prior and posterior.}
We compute visible structural support and infrared structural support as
\begin{gather}
\mathbf{S}_{v}
=
\mathrm{Norm}\!\left(|\nabla \mathbf{Y}_{v}|\right),
\qquad
\mathbf{S}_{r}
=
\max\!\left(
\mathrm{Norm}\!\left(|\nabla \mathbf{Y}_{r}|\right),
\mathrm{Norm}\!\left(
\mathrm{Std}\!\left(|\mathcal{B}(\mathbf{Y}_{r})|\right)
\right)
\right),
\end{gather}
where $\mathcal{B}(\cdot)$ is a band-pass operator and $\mathrm{Std}(\cdot)$ computes local standard deviation. 
The detail prior compares the two structural supports:
\begin{gather}
\boldsymbol{\pi}_{d}
=
p(z_{d}=1\mid \mathcal{P}_{d})
=
\frac{
\mathbf{S}_{r}
}{
\mathbf{S}_{r}+\mathbf{S}_{v}+\epsilon
},
\end{gather}
Since the structure prior alone cannot determine whether the current infrared detail is reliable for final rendering, PCDFusion further uses the context-updated state and local cross-modal observations to estimate the log-evidence score and posterior action weight:
\begin{gather}
\boldsymbol{\eta}_{d}
=
f^{d}_{\theta}
(
\mathbf{F},
\mathbf{Y}_{v},
\mathbf{Y}_{r},
\mathbf{Y}_{c},
\mathbf{S}_{v},
\mathbf{S}_{r}
)
,
\qquad
\mathbf{q}_{d}
=
\sigma\!\left(
\operatorname{logit}(\boldsymbol{\pi}_{d})+\boldsymbol{\eta}_{d}
\right).
\end{gather}

\noindent \textbf{Detail rendering.}
To avoid transferring low-frequency thermal background into the final appearance, PCDFusion predicts a bounded residual and keeps only its band-limited component before posterior-weighted rendering:
\begin{gather}
\mathbf{D}_{r}
=
s_{d}
\mathcal{B}\!\left(
\tanh\!\left(
d^{d}_{\theta}
(
\mathbf{F},
\mathbf{Y}_{v},
\mathbf{Y}_{r},
\mathbf{Y}_{c}
)
\right)
\right),
\qquad
\hat{\mathbf{Y}}
=
\mathbf{Y}_{c}
+
\mathbf{q}_{d}\odot \mathbf{D}_{r}.
\end{gather}
Here, $\mathbf{D}_{r}$ is the band-limited infrared residual, $s_{d}$ controls its scale, and $\hat{\mathbf{Y}}$ is the final fused luminance.

\subsection{Loss Functions}
\label{sec:objective}
PCDFusion is optimized with three losses for proposal recall, context posterior calibration, and final structural fidelity:
\begin{gather}
\mathcal{L}
=
\lambda_{\mathrm p}\mathcal{L}_{\mathrm p}
+
\lambda_{\mathrm c}\mathcal{L}_{\mathrm c}
+
\lambda_{\mathrm d}\mathcal{L}_{\mathrm d},
\end{gather}
where $\lambda_{\mathrm p}$, $\lambda_{\mathrm c}$, and $\lambda_{\mathrm d}$ weight the three loss terms, respectively.

\noindent \textbf{Proposal response recall.}
The proposal rendering $\mathbf{Y}_{p}$ should retain possible thermal responses without forcing them into the final rendering. 
We derive an infrared-salient mask $\mathbf{M}_{r}$ from $\mathbf{Y}_{r}$ and define a relaxed thermal target and one-sided recall loss as
\begin{gather}
\mathbf{T}_{r}
=
\max\!\left(
\mathbf{Y}_{v}
-
\boldsymbol{\pi}_{c}^{2}\odot
\mathrm{ReLU}(\mathbf{Y}_{v}-\mathbf{Y}_{r}),
\mathbf{Y}_{r}
\right),
\qquad
\mathcal{L}_{p}
=
\frac{
\left\|
\mathbf{M}_{r}
\odot
\mathrm{ReLU}(\mathbf{T}_{r}-\mathbf{Y}_{p})
\right\|_{1}
}{
\left\|\mathbf{M}_{r}\right\|_{1}+\epsilon
}.
\end{gather}

\noindent \textbf{Context posterior calibration.}
The context action weight should activate only where unreliable visible observation and infrared structure jointly support context update. 
We use a weak calibration target and a keep term for reliable visible regions:
\begin{gather}
\widetilde{\mathbf{q}}_{c}
=
\operatorname{sg}
\left(
\boldsymbol{\pi}_{c}\odot \mathbf{S}_{r}
\right),
\qquad
\mathcal{L}_{c}
=
\left\|
\mathbf{q}_{c}
-
\widetilde{\mathbf{q}}_{c}
\right\|_{1}
+
\lambda_{\mathrm{keep}}
\left\|
\operatorname{sg}(1-\boldsymbol{\pi}_{c})
\odot
\mathbf{q}_{c}
\right\|_{1},
\end{gather}
where $\operatorname{sg}(\cdot)$ stops gradients through the calibration target and the reliable-region mask.

\noindent \textbf{Final structural fidelity.}
The final rendering should preserve the strongest local structure without direct intensity forcing. 
We therefore match the gradient magnitude of $\hat{\mathbf{Y}}$ to the element-wise maximum gradient response:
\begin{gather}
\mathcal{L}_{d}
=
\left\|
|\nabla \hat{\mathbf{Y}}|
-
\max\!\left(
|\nabla \mathbf{Y}_{v}|,
|\nabla \mathbf{Y}_{r}|
\right)
\right\|_{1}.
\end{gather}

\begin{table*}[t]       
\centering
\caption{Quantitative fusion results on the MSRS, FMB, and M3FD datasets. The best results are highlighted in \textbf{bold}, and the second-best results are \underline{underlined}.}
\label{tab:main_results}
\setlength{\tabcolsep}{2pt}
\renewcommand{\arraystretch}{1.1}
\resizebox{\textwidth}{!}{%
\begin{tabular}{l|ccccc|ccccc|ccccc}
\toprule
\multirow{2}{*}{Method} 
& \multicolumn{5}{c|}{MSRS} 
& \multicolumn{5}{c|}{FMB} 
& \multicolumn{5}{c}{M3FD} \\
\cmidrule(lr){2-6} \cmidrule(lr){7-11} \cmidrule(lr){12-16}
& AG$\uparrow$ & SCD$\uparrow$ & VIF$\uparrow$ & Qabf$\uparrow$ & Qcb$\uparrow$
& AG$\uparrow$ & SCD$\uparrow$ & VIF$\uparrow$ & Qabf$\uparrow$ & Qcb$\uparrow$
& AG$\uparrow$ & SCD$\uparrow$ & VIF$\uparrow$ & Qabf$\uparrow$ & Qcb$\uparrow$ \\
\midrule
SwinFusion\venue{JAS'22} 
& 3.56 & \underline{1.69} & 1.03 & 0.65 & 0.58
& 4.05 & 1.56 & 1.03 & 0.66 & 0.46
& 4.61 & 1.56 & 0.90 & 0.61 & 0.46 \\

TarDAL\venue{CVPR'22} 
& 2.93 & 1.37 & 0.72 & 0.42 & 0.27
& 3.54 & 1.57 & 0.81 & 0.46 & 0.40
& 4.26 & 1.52 & 0.71 & 0.43 & 0.41 \\

CDDFuse\venue{CVPR'23} 
& 3.75 & 1.62 & 1.08 & 0.68 & 0.57
& 4.31 & 1.62 & 1.06 & 0.67 & 0.47
& 4.86 & 1.65 & 0.92 & 0.61 & 0.46 \\

Diff-IF\venue{Inf. Fusion'24} 
& 3.70 & 1.62 & \underline{1.10} & 0.69 & \underline{0.59}
& 4.05 & 1.37 & 1.05 & 0.66 & 0.46
& 4.57 & 1.32 & 0.89 & 0.59 & 0.45 \\

EMMA\venue{CVPR'24} 
& \underline{3.79} & 1.63 & 1.02 & 0.63 & 0.56
& \textbf{4.68} & 1.50 & 1.03 & 0.64 & 0.46
& \textbf{5.33} & 1.49 & 0.90 & 0.59 & 0.46 \\

DCEvo\venue{CVPR'25} 
& 3.72 & 1.66 & 1.07 & \underline{0.69} & 0.59
& 4.05 & 1.50 & 1.06 & \underline{0.69} & 0.47
& 4.62 & 1.49 & 0.92 & 0.63 & 0.47 \\

Deno-IF\venue{NIPS'25} 
& 3.11 & \textbf{1.79} & 0.90 & 0.56 & 0.54
& 3.19 & \underline{1.65} & 0.83 & 0.57 & 0.42
& 3.71 & \underline{1.68} & 0.74 & 0.53 & 0.42 \\

SAGE\venue{CVPR'25} 
& 3.16 & 1.42 & 0.76 & 0.54 & 0.50
& 3.78 & \textbf{1.71} & 0.97 & 0.64 & 0.45
& 4.25 & \textbf{1.69} & 0.86 & 0.58 & 0.45 \\

UPFusion\venue{AAAI'26} 
& 3.56 & 1.61 & 1.01 & 0.64 & 0.58
& 4.10 & 1.43 & \textbf{1.08} & 0.68 & \underline{0.47}
& 4.70 & 1.39 & \underline{0.95} & \underline{0.64} & \underline{0.47} \\

\rowcolor{gray!20}
PCDFusion\venue{Ours}
& \textbf{3.84} & 1.67 & \textbf{1.12} & \textbf{0.72} & \textbf{0.59}
& \underline{4.33} & 1.57 & \underline{1.08} & \textbf{0.71} & \textbf{0.49}
& \underline{5.04} & 1.58 & \textbf{0.95} & \textbf{0.67} & \textbf{0.50} \\
\bottomrule
\end{tabular}%
}
\vspace{0.8em}

\includegraphics[width=\textwidth]{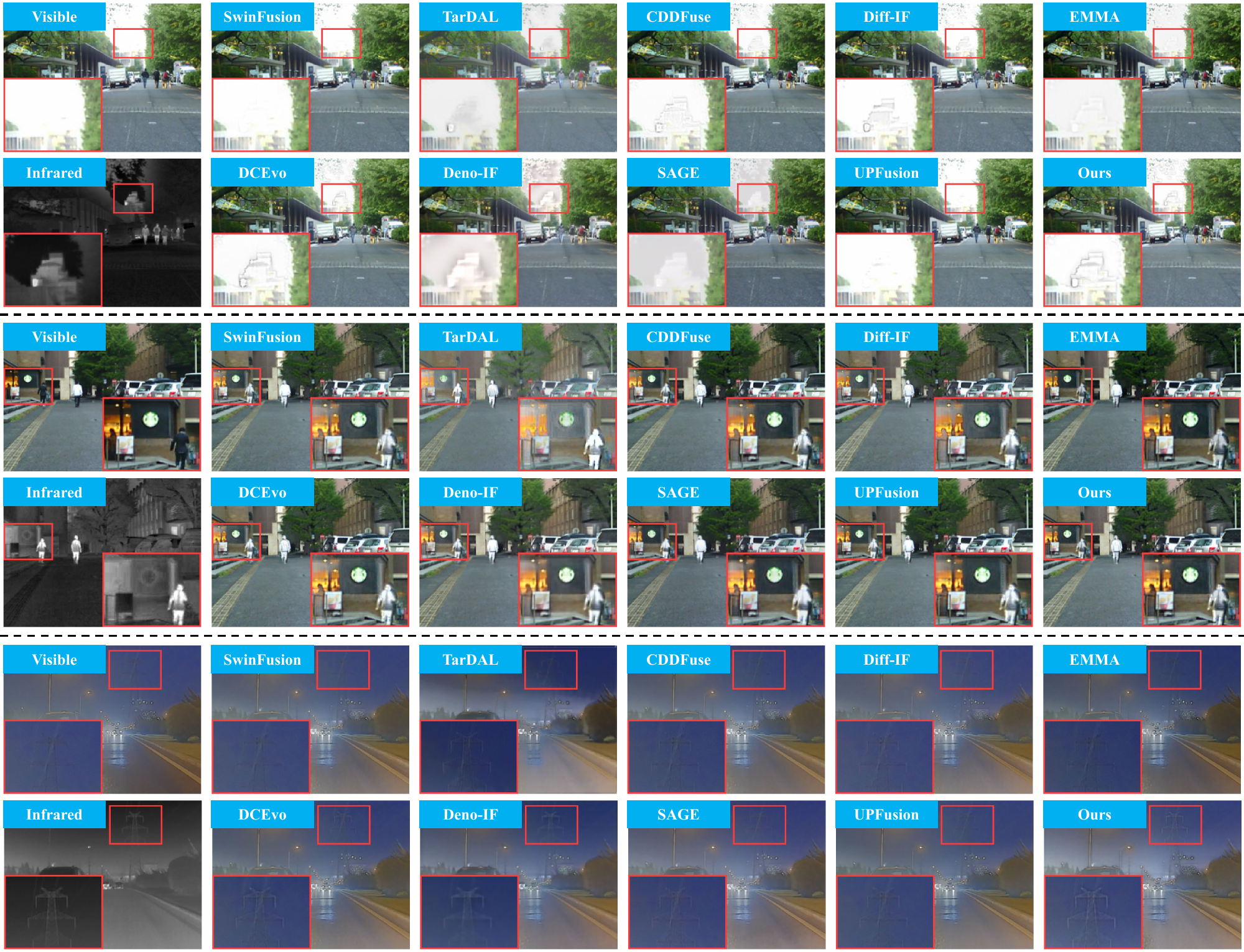}
\captionof{figure}{Visual comparisons of infrared-visible fusion results on MSRS and FMB dataset scenes.}
\label{fig:visual_msrs_fmb}
\end{table*}

\section{Experiments}
\label{sec:experiments}
\subsection{Settings and Details}
We evaluate PCDFusion on three public infrared-visible benchmarks: MSRS~\cite{piafusion}, FMB~\cite{segmif}, and M3FD~\cite{tardal}. 
The fusion model is trained only on the MSRS training split and directly evaluated on all datasets without dataset-specific fine-tuning. 
Following common IVIF evaluation protocols~\cite{ma2019survey,zhang2023visible}, we report AG, SCD~\cite{aslantas2015scd}, VIF~\cite{sheikh2006vif}, Qabf~\cite{xydeas2000qabf}, and Qcb~\cite{chen2009qcb}.
For downstream perception, we use SegFormerB2~\cite{segformer} for semantic segmentation and YOLOv10s~\cite{wang2024yolov10} for object detection. 
For each fusion method, the corresponding fused training images are used to fine-tune the task model from the same pretrained initialization under identical settings, and the fine-tuned model is then evaluated on the fused test images. 
We report mIoU on FMB and mAP$_{50:95}$ on M3FD. 
PCDFusion is implemented in PyTorch and trained for 200 epochs using AdamW with an initial learning rate of $2\times10^{-4}$, weight decay $10^{-4}$, batch size 16, crop size $256\times256$, and cosine learning-rate decay. 
During inference, the final RGB image is reconstructed by combining the fused luminance with the visible chrominance.

\begin{table*}[t]       
\centering
\caption{Downstream perception results on FMB semantic segmentation and M3FD object detection.}
\label{tab:down_results}
\setlength{\tabcolsep}{3pt}
\renewcommand{\arraystretch}{1.1}
\resizebox{\textwidth}{!}{%
\begin{tabular}{l|ccccccc|ccccccc}
\toprule
\multirow{2}{*}{Method}
& \multicolumn{7}{c|}{FMB Segmentation} 
& \multicolumn{7}{c}{M3FD Detection} \\
\cmidrule(lr){2-8} \cmidrule(lr){9-15}
& Side. & Light & Sign & Car & Truck & Bus & mIoU $\uparrow$
& Per. & Car & Bus & Motor. & Lamp & Truck & mAP$_{50:95}$ $\uparrow$ \\
\midrule
SwinFusion\venue{JAS'22}         
& 48.98 & 43.22 & 80.68 & 84.23 & \underline{48.77} & 57.57 & 58.33
& 48.90 & 68.13 & 71.31 & 40.82 & 42.03 & 58.00 & 54.86 \\

TarDAL\venue{CVPR'22}            
& 50.05 & 42.30 & 81.20 & 81.85 & 35.82 & 26.36 & 55.59
& 48.80 & 66.51 & 69.55 & 34.00 & 38.87 & 56.38 & 52.35 \\

CDDFuse\venue{CVPR'23}           
& \underline{50.97} & 40.85 & 81.80 & 82.90 & 22.14 & 55.70 & 56.66
& 48.79 & 68.44 & \underline{71.80} & 38.00 & 43.44 & 54.39 & 54.15 \\

Diff-IF\venue{Inf. Fusion'24}    
& 43.81 & 45.24 & \underline{82.04} & 83.57 & 45.22 & \textbf{75.99} & \underline{58.98}
& 49.77 & 67.43 & 70.08 & 38.47 & 39.82 & \underline{58.68} & 54.04 \\

EMMA\venue{CVPR'24}              
& 49.32 & \textbf{49.42} & 81.24 & 82.58 & 42.99 & 59.41 & 58.47
& 48.35 & 68.03 & 71.02 & 38.26 & 43.70 & 57.20 & 54.43 \\

DCEvo\venue{CVPR'25}             
& 48.01 & \underline{46.98} & 81.01 & \textbf{84.55} & 43.67 & 49.08 & 57.73
& \textbf{50.05} & 68.25 & 71.60 & \underline{41.20} & 43.86 & \textbf{59.78} & \underline{55.79} \\

Deno-IF\venue{NIPS'25}           
& 47.39 & 40.84 & 80.57 & 82.25 & 40.13 & 34.40 & 55.45
& \underline{50.03} & \underline{68.68} & 70.37 & 41.06 & 43.16 & 57.61 & 55.15 \\

SAGE\venue{CVPR'25}              
& \textbf{51.65} & 42.90 & \textbf{82.37} & 82.28 & 41.08 & 31.51 & 56.33
& 49.59 & 68.42 & 68.91 & 39.15 & 43.22 & 56.39 & 54.28 \\

UPFusion\venue{AAAI'26}          
& 43.86 & 19.68 & 77.97 & 81.92 & 39.27 & 58.29 & 55.36
& 48.72 & 68.52 & 71.28 & 38.57 & \textbf{45.94} & 58.04 & 55.18 \\

\rowcolor{gray!20}
PCDFusion\venue{Ours}                            
& 50.79 & 45.89 & 81.31 & \underline{84.44} & \textbf{51.54} & \underline{71.19} & \textbf{60.05}
& 48.95 & \textbf{69.00} & \textbf{72.60} & \textbf{42.89} & \underline{44.53} & 57.93 & \textbf{55.98} \\
\bottomrule
\end{tabular}%
}
\vspace{0.8em}

\includegraphics[width=\textwidth]{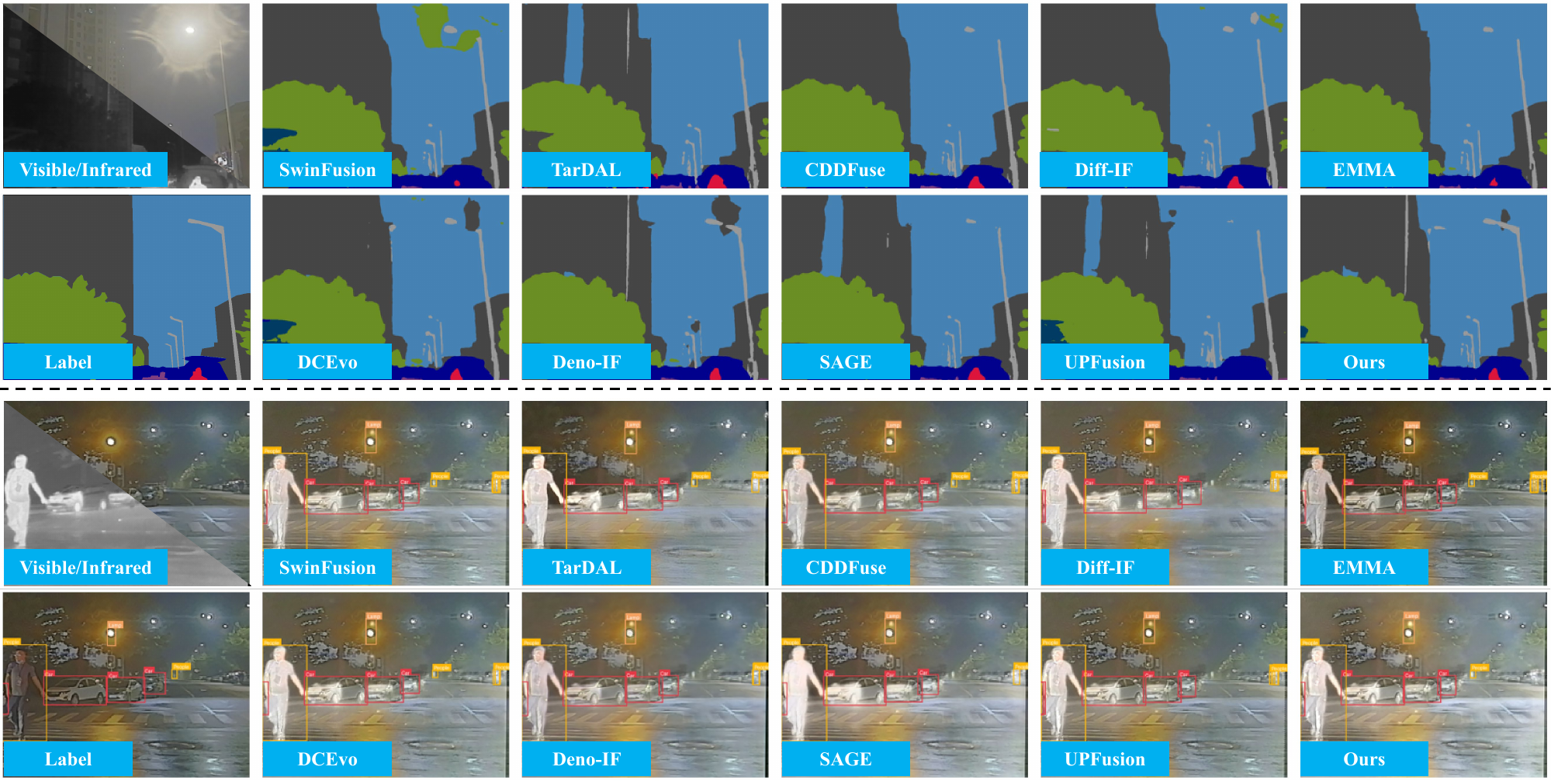}
\captionof{figure}{Visualization results on downstream perception tasks.}
\label{fig:visual_downstream}
\end{table*}

\subsection{Results of Infrared-Visible Image Fusion}
We compare PCDFusion with recent representative IVIF methods, including SwinFusion~\cite{swinfusion}, TarDAL~\cite{tardal}, CDDFuse~\cite{cddfuse}, Diff-IF~\cite{diffif}, EMMA~\cite{emma}, DCEvo~\cite{dcevo}, Deno-IF~\cite{denoif}, SAGE~\cite{sage}, and UPFusion~\cite{upfusion}.

\noindent\textbf{Quantitative comparisons.}
\cref{tab:main_results} reports fusion results on MSRS, FMB, and M3FD. 
PCDFusion achieves the best overall trade-off, consistently leading in VIF, Qabf, and Qcb across the three datasets. 
Although some methods obtain higher AG or SCD on individual datasets, their gains are less consistent across other quality measures. 
This indicates that PCDFusion provides more stable fusion quality under heterogeneous modality reliability.

\noindent\textbf{Qualitative comparisons.}
\Cref{fig:visual_msrs_fmb} shows representative MSRS and FMB scenes with local zoom-ins. 
In the first scene, PCDFusion restores clearer building structures in the saturated visible region without contaminating the surrounding sky. 
In the second scene, the visible image remains reliable, while the infrared observation fails around the glass region. Competing methods tend to inject unreliable thermal responses or disturb the visible appearance, whereas PCDFusion suppresses the invalid infrared response and preserves the original structure. 
In the third scene, haze and low contrast make thin structures difficult to preserve. 
PCDFusion maintains clearer utility-pole structures while avoiding excessive thermal amplification in the background. 
These results show that PCDFusion can both recover useful thermal structures under visible failure and reject unreliable infrared responses when visible appearance is reliable.

\begin{table*}[t]
\centering
\begin{minipage}[t]{0.465\textwidth}
\centering
\caption{Ablation of PCDFusion stages.}
\label{tab:ablation_component}
\setlength{\tabcolsep}{3pt}
\renewcommand{\arraystretch}{1.1}
\resizebox{\linewidth}{!}{%
\begin{tabular}{l|ccccc|cc}
\toprule
Method & AG$\uparrow$ & SCD$\uparrow$ & VIF$\uparrow$ & Qabf$\uparrow$ & Qcb$\uparrow$ & mIoU$\uparrow$ & mAP$\uparrow$ \\
\midrule
w/o P
& 3.810 & 1.594 & \underline{1.116} & 0.723 & \underline{0.592} & 58.44 & 55.18 \\
w/o C
& 3.825 & \textbf{1.716} & 1.107 & 0.719 & 0.582 & 58.86 & 54.88 \\
w/o D
& \textbf{3.846} & 1.618 & 1.110 & \textbf{0.723} & 0.592 & \underline{58.89} & \textbf{56.13} \\
\rowcolor{gray!20}
Full
& \underline{3.839} & \underline{1.671} & \textbf{1.117} & \underline{0.723} & \textbf{0.593} & \textbf{60.05} & \underline{55.98} \\
\bottomrule
\end{tabular}%
}
\end{minipage}
\hfill
\begin{minipage}[t]{0.515\textwidth}
\centering
\caption{Ablation of posterior estimation forms.}
\label{tab:ablation_posterior_form}
\setlength{\tabcolsep}{3pt}
\renewcommand{\arraystretch}{1.1}
\resizebox{\linewidth}{!}{%
\begin{tabular}{l|ccccc|cc}
\toprule
Method 
& AG$\uparrow$ & SCD$\uparrow$ & VIF$\uparrow$ & Qabf$\uparrow$ & Qcb$\uparrow$
& mIoU$\uparrow$ & mAP$\uparrow$ \\
\midrule
only Prior
& 3.789 & 1.666 & \textbf{1.123} & \textbf{0.728} & \textbf{0.597}
& \underline{59.09} & 55.56 \\

only Evidence 
& \textbf{3.885} & \textbf{1.696} & 1.104 & 0.722 & 0.589
& 59.04 & 55.38 \\

Product
& 3.835 & 1.596 & 1.102 & 0.722 & 0.588
& 58.89 & \underline{55.76} \\

\rowcolor{gray!20}
Full
& \underline{3.839} & \underline{1.671} & \underline{1.117} & \underline{0.723} & \underline{0.593}
& \textbf{60.05} & \textbf{55.98} \\
\bottomrule
\end{tabular}%
}
\end{minipage}
\end{table*}

\begin{figure}[t]
    \centering
    \includegraphics[width=\textwidth]{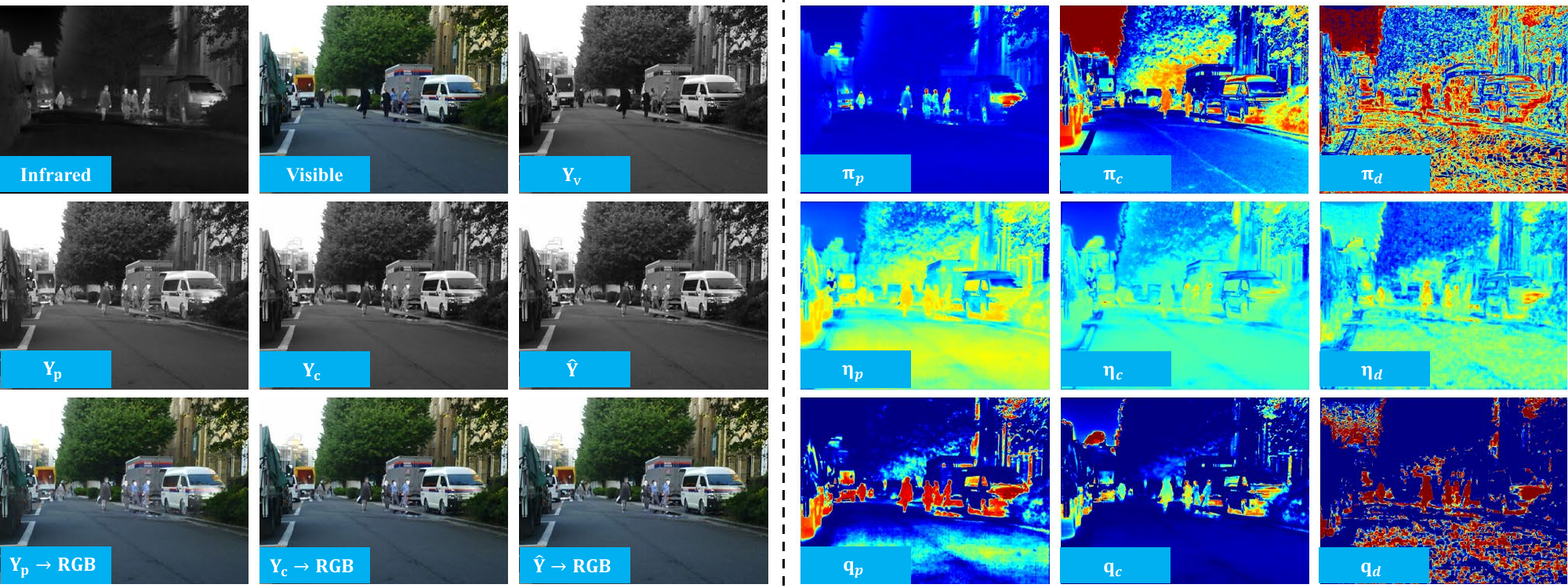}
    \caption{
    Left: luminance and RGB outputs of the proposal, context, and detail rendering stages. 
    Right: priors, learned log-evidence scores, and posterior action weights of the three Bayesian processes.
    }
    \label{fig:visual_stage}
\end{figure}

\begin{figure}[t]
\centering
\begin{minipage}[t]{0.48\linewidth}
    \centering
    \includegraphics[width=\linewidth]{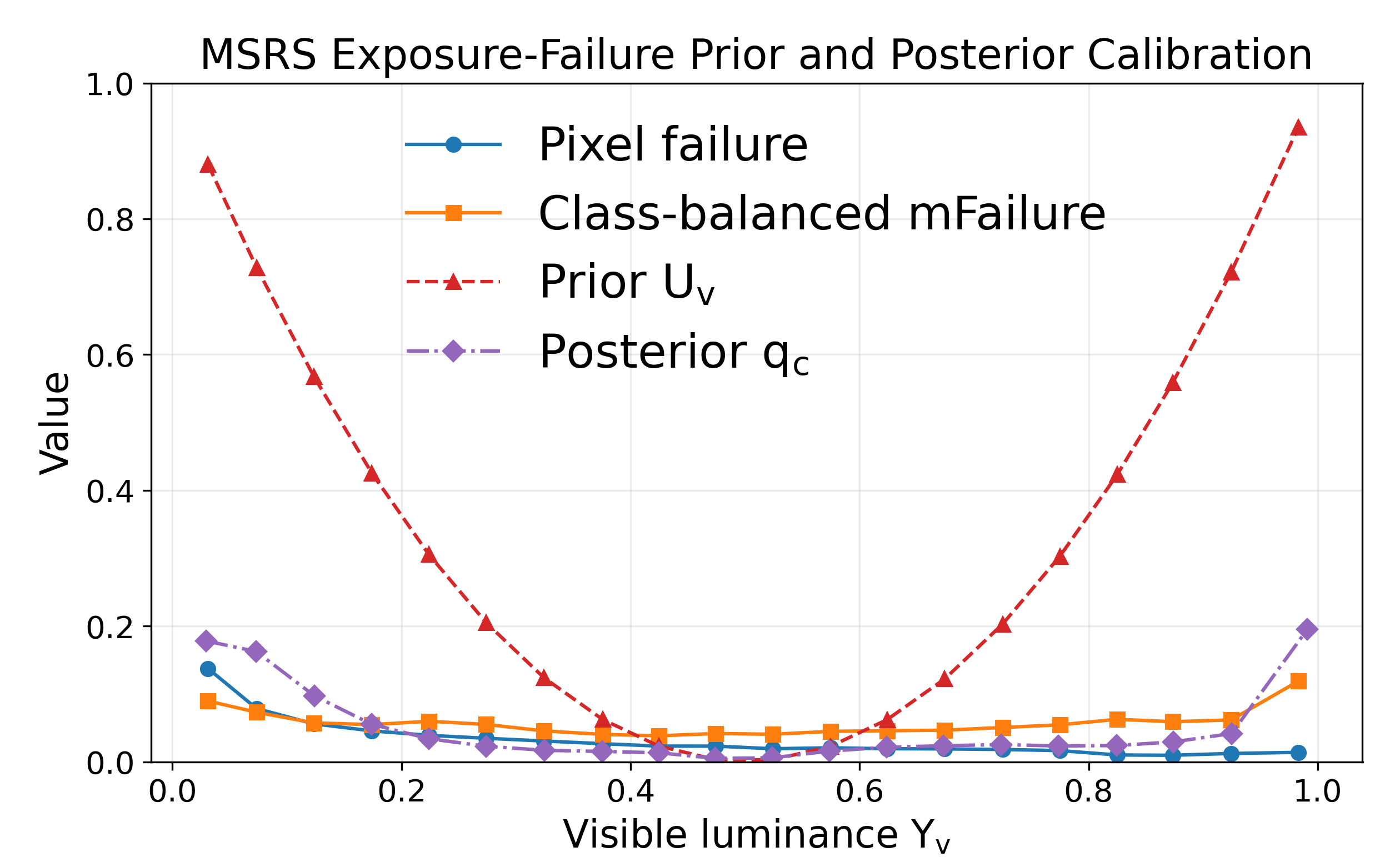}
\end{minipage}
\hfill
\begin{minipage}[t]{0.48\linewidth}
    \centering
    \includegraphics[width=\linewidth]{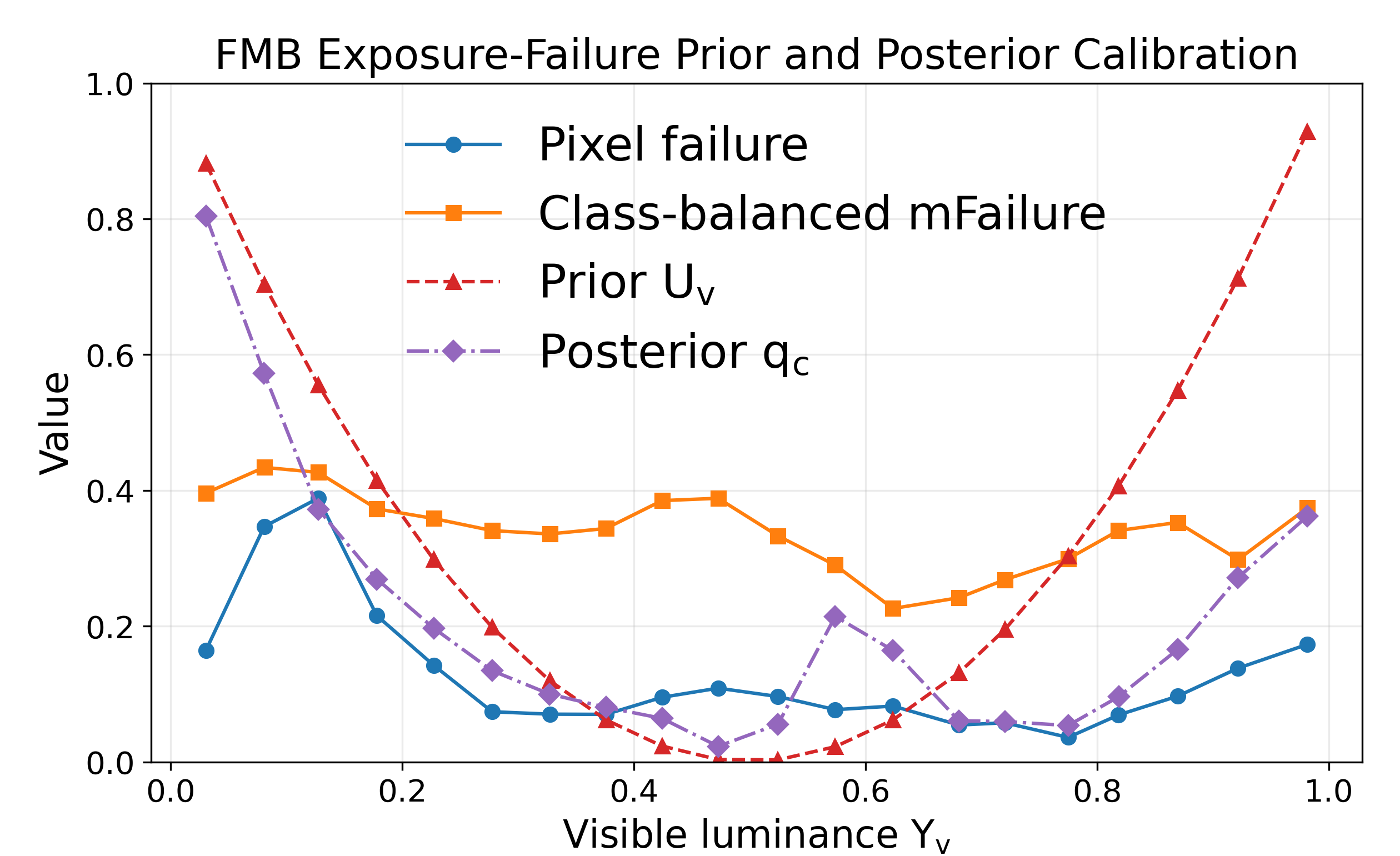}
\end{minipage}
\caption{Calibration of the exposure-failure prior and posterior context update weight.}
\label{fig:plot_prior}
\end{figure}

\begin{table*}[t]
\centering
\begin{minipage}[t]{0.565\textwidth}
\centering
\caption{Ablation of context prior estimates.}
\label{tab:ablation_context_prior}
\setlength{\tabcolsep}{3pt}
\renewcommand{\arraystretch}{1.1}
\resizebox{\linewidth}{!}{%
\begin{tabular}{l|ccccc|cc}
\toprule
Priors
& AG$\uparrow$ & SCD$\uparrow$ & VIF$\uparrow$ & Qabf$\uparrow$ & Qcb$\uparrow$
& mIoU$\uparrow$ & mAP$\uparrow$ \\
\midrule
$1-\mathbf{Y}_{v}$
& 3.839 & \textbf{1.693} & \underline{1.111} & \textbf{0.723} & \underline{0.589}
& 58.88 & \underline{55.53} \\

$\mathbf{Y}_{v}$
& \textbf{3.875} & \underline{1.692} & 1.106 & 0.722 & 0.588
& \underline{59.19} & 55.46 \\

$\left|2\mathbf{Y}_{v}-1\right|$
& \underline{3.863} & 1.675 & 1.107 & 0.721 & 0.587
& 58.94 & 55.50 \\

\rowcolor{gray!20}
$\left(2\mathbf{Y}_{v}-1\right)^2$
& 3.839 & 1.671 & \textbf{1.117} & \textbf{0.723} & \textbf{0.593}
& \textbf{60.05} & \textbf{55.98} \\
\bottomrule
\end{tabular}%
}
\end{minipage}
\hfill
\begin{minipage}[t]{0.415\textwidth}
\centering
\caption{Ablation of objectives.}
\label{tab:ablation_objective}
\setlength{\tabcolsep}{3pt}
\renewcommand{\arraystretch}{1.1}
\resizebox{\linewidth}{!}{%
\begin{tabular}{l|ccccc|cc}
\toprule
Method & AG$\uparrow$ & SCD$\uparrow$ & VIF$\uparrow$ & Qabf$\uparrow$ & Qcb$\uparrow$ & mIoU$\uparrow$ & mAP$\uparrow$ \\
\midrule
w/o $\mathcal{L}_{p}$
& \underline{3.868} & 1.658 & 1.105 & 0.721 & 0.589 & \underline{59.02} & \underline{56.00} \\

w/o $\mathcal{L}_{c}$
& \textbf{3.872} & \underline{1.678} & \underline{1.111} & \underline{0.723} & \textbf{0.594} & 58.90 & 55.58 \\

w/o $\mathrm{keep}$
& 3.862 & 1.675 & 1.107 & 0.722 & 0.586 & 58.79 & \textbf{56.01} \\

w/o $\mathcal{L}_{d}$
& 3.798 & 1.070 & 1.010 & 0.178 & 0.332 & 41.01 & 23.83 \\

Final Sal.
& 3.852 & \textbf{1.735} & 1.098 & 0.720 & 0.582 & 58.56 & 55.15 \\

\rowcolor{gray!20}
Full
& 3.839 & 1.671 & \textbf{1.117} & \textbf{0.723} & \underline{0.593} & \textbf{60.05} & 55.98 \\
\bottomrule
\end{tabular}%
}
\end{minipage}
\end{table*}

\subsection{Results of Downstream Tasks}
\cref{tab:down_results} reports downstream results on FMB segmentation and M3FD detection. 
PCDFusion achieves the best aggregate mIoU and mAP$_{50:95}$, although no method ranks first for every class. Its $60.05\%$ mIoU exceeds Diff-IF ($58.98\%$) by $1.07$ percentage points, and its $55.98\%$ mAP$_{50:95}$ exceeds DCEvo ($55.79\%$) by $0.19$ percentage points.

\cref{fig:visual_downstream} shows representative downstream visualizations. 
For segmentation, PCDFusion produces a more complete pole shape, while competing methods often fragment or miss thin structures. 
For detection, PCDFusion avoids falsely detecting the bright region on the far right and maintains reliable localization of valid targets. 
These results indicate cleaner structure preservation and less background-induced task interference.

\subsection{Discussions}
\label{sec:ablation}

To analyze the effectiveness of PCDFusion, we conduct ablation studies on MSRS.

\noindent\textbf{Effectiveness of different rendering stages.}
\cref{tab:ablation_component} and the stage outputs in the left of \cref{fig:visual_stage} show that the three stages perform complementary rendering functions. 
Removing P causes the most evident degradation, confirming that early preservation of positive thermal responses is necessary. 
Removing C weakens reliable reconstruction despite increasing SCD, indicating that context recovery cannot be driven only by cross-modal variation. 
Removing D reduces the final balance, showing that band-limited thermal detail still requires selective rendering. 
$\mathbf{Y}_{p}$ incorporates infrared targets such as pedestrians and vehicles, but still misses details in the over-exposed building region; $\mathbf{Y}_{c}$ restores the building structure in the visible-failure area; and $\hat{\mathbf{Y}}$ further sharpens the recovered details.

\noindent\textbf{Effectiveness of Bayesian priors and posteriors.}
\cref{tab:ablation_posterior_form,tab:ablation_context_prior} analyze the Bayesian design. 
Using only priors yields strong image-level fidelity, confirming that the physical cues are informative; using only learned evidence improves AG and SCD but weakens downstream performance, suggesting over-amplified local variations. 
The product form is less stable than the normalized posterior, while the full update gives the best downstream results with balanced fusion quality. 
For context modeling, one-sided or linear exposure priors are less effective than the adopted symmetric exposure-failure prior, especially on VIF, Qcb, mIoU, and mAP. 
The Bayesian maps in the right of \cref{fig:visual_stage} show that the posteriors refine their priors: $\mathbf{q}_{p}$ strengthens reliable thermal-response regions, $\mathbf{q}_{c}$ restricts context update to unreliable visible areas, and $\mathbf{q}_{d}$ localizes final detail compensation. 
In \cref{fig:plot_prior}, $\mathbf{q}_{c}$ also follows the empirical exposure-failure trends on MSRS and FMB, indicating good posterior calibration across datasets.

\noindent\textbf{Effectiveness of loss functions.}
\cref{tab:ablation_objective} validates the objective design. 
Removing $\mathcal{L}_{p}$ weakens proposal recall and reduces downstream performance, showing that early thermal-response preservation requires explicit supervision. 
Removing $\mathcal{L}_{c}$ or the keep term can improve some image-level scores, but lowers mIoU or mAP, indicating that context posterior calibration is necessary for stable task utility. 
Removing $\mathcal{L}_{d}$ causes a severe collapse in Qabf, Qcb, mIoU, and mAP, confirming that final structural fidelity is indispensable. 
Directly applying saliency supervision to the final output increases SCD, but weakens VIF, Qabf, Qcb, and downstream results, supporting our choice to supervise thermal responses at the proposal stage rather than force them into the final rendering.

\begin{table*}[t]
\centering
\begin{minipage}[t]{0.6\textwidth}
\centering
\caption{FMB mIoU under input corruption. Slash-separated values follow severities 1/2/3.}
\label{tab:ablation_corruption}
\setlength{\tabcolsep}{3pt}
\renewcommand{\arraystretch}{1.08}
\resizebox{\linewidth}{!}{%
\begin{tabular}{l|c|ccc}
\toprule
Corruption & Severity & UPFusion & SAGE & PCDFusion \\
\midrule
Visible dark
& 1/2/3
& 55.93 / 51.75 / 47.37
& \underline{56.72} / \underline{53.79} / \underline{51.02}
& \textbf{59.75} / \textbf{57.09} / \textbf{52.14} \\

Visible bright
& 1/2/3
& 54.97 / 50.61 / 45.98
& \underline{56.86} / \underline{55.33} / \textbf{51.28}
& \textbf{59.84} / \textbf{56.07} / \underline{50.28} \\

Infrared only
& 1/2/3
& 54.62 / 53.76 / 53.10
& \underline{55.94} / \underline{55.43} / \underline{54.55}
& \textbf{58.06} / \textbf{57.44} / \textbf{56.54} \\
\midrule
Both (VIS dark)
& 1/2/3
& 54.05 / 50.41 / 42.90
& \underline{56.00} / \underline{53.38} / \textbf{50.06}
& \textbf{57.66} / \textbf{53.45} / \underline{44.23} \\

Both (VIS bright)
& 1/2/3
& 53.92 / 48.75 / 43.32
& \underline{56.23} / \textbf{54.27} / \textbf{47.85}
& \textbf{58.46} / \underline{53.94} / \underline{45.83} \\
\bottomrule
\end{tabular}%
}
\end{minipage}
\hfill
\begin{minipage}[t]{0.37\textwidth}
\centering
\caption{Ablation of rendering order and joint decisions.}
\label{tab:ablation_order}
\setlength{\tabcolsep}{3pt}
\renewcommand{\arraystretch}{1.08}
\resizebox{\linewidth}{!}{%
\begin{tabular}{l|ccccc}
\toprule
Order
& AG$\uparrow$ & VIF$\uparrow$ & Qabf$\uparrow$
& mIoU$\uparrow$ & mAP$\uparrow$ \\
\midrule
\rowcolor{gray!20}
P--C--D 
& \textbf{3.839} & \textbf{1.117} & \textbf{0.723}
& \textbf{60.05} & \textbf{55.98} \\

C--P--D
& 3.820 & 1.091 & 0.715
& 56.30 & 54.86 \\

P--D--C
& 3.792 & \underline{1.110} & \underline{0.720}
& 56.24 & \underline{55.68} \\

Parallel
& \underline{3.824} & 1.107 & 0.718
& \underline{56.49} & 55.33 \\
\bottomrule
\end{tabular}%
}
\end{minipage}
\end{table*}

\noindent\textbf{Robustness under input corruption.}
\Cref{tab:ablation_corruption} evaluates PCDFusion under both single- and dual-modal corruption.
Across the 15 corruption settings, PCDFusion ranks first in 11 and second in the remaining four, while consistently outperforming UPFusion.
When only one modality is degraded, PCDFusion achieves the highest mIoU in eight of nine settings, indicating that the posterior rendering mechanism can exploit reliable complementary information from the other modality.
Under joint corruption, it remains the best method for both severity-1 settings and the severity-2 VIS-dark setting, and maintains higher mIoU than UPFusion in all six cases.
Although SAGE performs better under several severe joint corruptions, PCDFusion exhibits a more gradual performance decrease as the corruption strength increases.
These results demonstrate that the physical priors and observation-dependent posterior weights effectively adapt the contribution of each modality to its local reliability, improving robustness over fixed fusion strategies.

\noindent\textbf{Sequential versus joint rendering decisions.}
\Cref{tab:ablation_order} evaluates whether the three rendering decisions should be performed sequentially or jointly.
The proposed P--C--D sequence achieves the best result on every fusion and downstream metric.
Compared with the strongest alternative for each metric, it improves AG, VIF, Qabf, mIoU, and mAP by 0.015, 0.007, 0.003, 3.56, and 0.30, respectively.
In particular, placing context update before proposal rendering reduces mIoU from 60.05 to 56.30, while applying detail rendering before context update reduces it to 56.24.
The parallel joint-gating variant also reaches only 56.49 mIoU, showing that the improvement cannot be explained merely by introducing three rendering decisions.
Instead, the stages benefit from their ordered dependency: proposal rendering first preserves candidate thermal responses, context rendering then corrects the visible representation using this proposal, and detail rendering finally injects structural residuals into the corrected representation.
The consistent gains across image-level quality and downstream performance validate the proposed sequential posterior-rendering design.

\noindent\textbf{Limitations.}
PCDFusion still has limitations in low-luminance and infrared-unreliable regions. 
As shown in \cref{fig:teaser}(b), although it reduces the influence of unreliable infrared responses, the tree branches remain slightly blurred. 
In the second scene of \cref{fig:visual_msrs_fmb}, the glass region is largely invalid in infrared, but minor thermal interference is still transferred into the fused result. 
A possible reason is that the context posterior in \cref{fig:plot_prior} remains relatively high in low-luminance ranges, which helps recover degraded visible regions but can also make the model over-permissive when infrared responses are unreliable. 
Future work will improve posterior calibration with stronger structure-preservation constraints and more explicit infrared reliability estimation.

\section{Conclusion}
\label{sec:conclusion}
In this paper, we presented PCDFusion, a visible-anchored posterior rendering framework for IVIF. 
PCDFusion formulates fusion as local reliability-conditioned rendering from heterogeneous visible and thermal observations. 
It uses proposal, context, and detail rendering to preserve positive thermal responses, update exposure-degraded visible regions, and inject band-limited infrared structures. 
By combining physical priors with learned log-evidence scores, PCDFusion selectively uses thermal responses while preserving reliable visible appearance. 
Experiments on public benchmarks and downstream tasks show improved fusion quality, target preservation, and task utility. 
Nevertheless, low-luminance regions and infrared-unreliable cases remain challenging, especially when weak visible structures should be preserved while invalid thermal responses should be suppressed. 
Future work will explore stronger posterior calibration and more explicit infrared reliability estimation.

\section{Acknowledgments}
This work was supported by the National NaturalScience Foundation of China ( 62671049 and  62331006), and Beijing Natural Science Foundation (L253009).

{
	\bibliographystyle{nips}
	\bibliography{reference}
}

\clearpage
\appendix
\begin{center}
    {\Large\bfseries Supplementary Material}
\end{center}

\section{Implementation Details}
\label{app:implementation_details}

\noindent\textbf{Network Implementation.}
PCDFusion follows the visible-anchored feature flow described in the main paper. 
The visible branch uses a single-scale Swin Transformer stream with patch size $8$, embedding dimension $192$, and depth $4$. 
The four Swin blocks are divided into two stages. 
The token resolution is kept at $H/8\times W/8$ throughout the visible backbone.

The infrared branch is implemented as a lightweight convolutional encoder with base width $24$. 
It extracts multi-scale infrared features at resolutions $H/2$, $H/4$, $H/8$, and $H/16$, with channel widths $48$, $96$, $192$, and $192$, respectively. 
All infrared features are projected to the visible token dimension and concatenated as infrared tokens $\mathbf{T}_{r}$. 
Let $\mathbf{T}_{v}$ denote the visible tokens. 
The infrared-to-visible injection is implemented by cross-attention:
\begin{gather}
\mathbf{Q}=\mathbf{T}_{v}\mathbf{W}_{Q},\qquad
\mathbf{K}=\mathbf{T}_{r}\mathbf{W}_{K},\qquad
\mathbf{V}=\mathbf{T}_{r}\mathbf{W}_{V},\\
\widetilde{\mathbf{T}}_{v}
=
\mathbf{T}_{v}
+
\operatorname{Softmax}
\left(
\frac{\mathbf{Q}\mathbf{K}^{\top}}{\sqrt{d}}
\right)
\mathbf{V}.
\end{gather}
Here, visible tokens provide queries, infrared tokens provide keys and values, and $d$ is the attention dimension.

After cross-modal injection, the visible token stream is unembedded back to the image resolution. 
The recovered visible-anchored feature is concatenated with the visible image and the infrared observation, and then mapped to a full-resolution hidden feature with width $48$. 
The proposal, context, and detail rendering heads are applied on this hidden feature to predict the posterior weights and rendering terms used in the main paper.

\begin{table*}[h!]
\centering
\caption{Probabilistic roles and their realizations in PCDFusion.}
\label{tab:app_bayesian_roles}
\setlength{\tabcolsep}{4pt}
\renewcommand{\arraystretch}{1.12}
\resizebox{\textwidth}{!}{%
\begin{tabular}{l|l|l|l}
\toprule
Role & Probabilistic operation & PCDFusion realization & Consequence \\
\midrule
Decision
& Bernoulli rendering action $z_k$
& Explicit proposal/context/detail action semantics
& Defined local inference target \\

Prior
& $p(z_k=1\mid\mathcal{P}_k)$
& Analytic IR-response, exposure, or structure cue
& Physical zero-evidence reference \\

Evidence
& Relative support in log Bayes-factor coordinates
& Neural score $f_{\theta_k}(\mathcal{O}_k,\mathcal{P}_k)$
& Amortized log-evidence correction \\

Update
& Prior odds $\times$ evidence factor
& $\mathbf{q}_k=\sigma(\operatorname{logit}(\boldsymbol{\pi}_k)+\boldsymbol{\eta}_k)$
& Normalized action posterior \\

Rendering
& Expectation over two actions
& $(1-\mathbf{q}_k)\mathcal{R}_{k,0}+\mathbf{q}_k\mathcal{R}_{k,1}$
& Posterior-weighted fusion \\

Learning
& Shared estimator of observation evidence
& End-to-end optimization of $\theta_k$ through rendering losses
& Amortized across pixels and images \\
\bottomrule
\end{tabular}%
}
\end{table*}

\Cref{tab:app_bayesian_roles} summarizes how the probabilistic quantities map to the implemented rendering pipeline. 
The binary variable specifies the local action being inferred, the analytic cue provides its physical prior, and the neural estimator supplies an observation-dependent correction in log-odds coordinates. 
The normalized posterior then mixes the base and action-conditioned outcomes, so all three rendering stages share the same inference structure while using different physical cues and outcomes.

\noindent\textbf{RGB Reconstruction.}
PCDFusion predicts the fused luminance $\hat{\mathbf{Y}}$. 
Given the visible RGB image 
$\mathbf{I}_{\mathrm v}=(\mathbf{R}_{\mathrm v},\mathbf{G}_{\mathrm v},\mathbf{B}_{\mathrm v})$, 
we first compute its luminance and chrominance channels by
\begin{gather}
\mathbf{Y}_{\mathrm v}
=
0.299\mathbf{R}_{\mathrm v}
+
0.587\mathbf{G}_{\mathrm v}
+
0.114\mathbf{B}_{\mathrm v},
\\
\mathbf{Cb}_{\mathrm v}
=
-0.168736\mathbf{R}_{\mathrm v}
-
0.331264\mathbf{G}_{\mathrm v}
+
0.5\mathbf{B}_{\mathrm v}
+
0.5,
\\
\mathbf{Cr}_{\mathrm v}
=
0.5\mathbf{R}_{\mathrm v}
-
0.418688\mathbf{G}_{\mathrm v}
-
0.081312\mathbf{B}_{\mathrm v}
+
0.5.
\end{gather}
The final RGB image is reconstructed by replacing the visible luminance with the predicted fused luminance $\hat{\mathbf{Y}}$ while preserving the visible chrominance:
\begin{gather}
\hat{\mathbf{R}}
=
\hat{\mathbf{Y}}
+
1.402(\mathbf{Cr}_{\mathrm v}-0.5),
\\
\hat{\mathbf{G}}
=
\hat{\mathbf{Y}}
-
0.344136(\mathbf{Cb}_{\mathrm v}-0.5)
-
0.714136(\mathbf{Cr}_{\mathrm v}-0.5),
\\
\hat{\mathbf{B}}
=
\hat{\mathbf{Y}}
+
1.772(\mathbf{Cb}_{\mathrm v}-0.5),
\\
\hat{\mathbf{I}}_{\mathrm{rgb}}
=
\operatorname{clip}
\left(
(\hat{\mathbf{R}},\hat{\mathbf{G}},\hat{\mathbf{B}}),
0,1
\right).
\end{gather}

\section{Experimental Details}
\label{app:experimental_protocol}
\noindent\textbf{Computational Environment.}
\label{app:compute_environment}
All training and inference experiments are conducted on a server with one NVIDIA RTX 4090 GPU and two Intel Xeon Platinum 8481C CPUs. 
Each CPU has $56$ physical cores and supports $2$ threads per core, resulting in $224$ logical CPU threads in total. 
Under the training setting reported in the main paper, PCDFusion uses $18{,}206$ MiB GPU memory and takes about $18$ seconds per epoch on the MSRS training split. 
For single-image inference on MSRS, the GPU memory usage is about $1{,}696$ MiB.

\noindent\textbf{Datasets and Training Split.}
\label{app:datasets}
PCDFusion is trained only on the MSRS training split, which contains $1{,}083$ aligned visible-infrared image pairs. 
For image-level fusion evaluation, we use the MSRS test split with $361$ pairs, the FMB test split with $280$ pairs, and $300$ pairs from M3FD. 
The fusion model is directly applied to all test sets without dataset-specific fine-tuning.

For downstream evaluation, we follow the task setting of each benchmark. 
For FMB semantic segmentation, we use $1{,}220$ training pairs and $280$ testing pairs. 
For M3FD object detection, the detector is trained on M3FD images outside the $300$ fusion-evaluation pairs and evaluated on the fused results of these $300$ pairs. 
Semantic segmentation labels and object detection annotations are used only to train and evaluate downstream task models, not to train PCDFusion.

\noindent\textbf{External Assets and Public Repositories.}
\label{app:external_assets}
We use public datasets, official baseline implementations, public metric code, and publicly released downstream checkpoints when available. 
\Cref{tab:app_external_assets} lists the external assets used in our experiments. 
All assets are used only for academic evaluation, following the terms specified by their corresponding repositories.

\begin{table*}[t]
\centering
\caption{External assets used in this work. ``Not specified'' means that no explicit repository-level license is shown on the public repository page.}
\label{tab:app_external_assets}
\setlength{\tabcolsep}{3pt}
\renewcommand{\arraystretch}{1.08}
\resizebox{\textwidth}{!}{%
\begin{tabular}{l|l|l|l}
\toprule
Asset & Type & Repository / Source & License / Terms \\
\midrule
SwinFusion\venue{JAS'22} 
& Baseline implementation 
& \url{https://github.com/Linfeng-Tang/SwinFusion.git} 
& Not specified \\

TarDAL\venue{CVPR'22} 
& Baseline implementation / M3FD source 
& \url{https://github.com/JinyuanLiu-CV/TarDAL.git} 
& GPL-3.0 \\

CDDFuse\venue{CVPR'23} 
& Baseline implementation 
& \url{https://github.com/Zhaozixiang1228/MMIF-CDDFuse.git} 
& Not specified \\

Diff-IF\venue{Inf. Fusion'24} 
& Baseline implementation 
& \url{https://github.com/XunpengYi/Diff-IF.git} 
& MIT \\

EMMA\venue{CVPR'24} 
& Baseline implementation 
& \url{https://github.com/Zhaozixiang1228/MMIF-EMMA.git} 
& Not specified \\

DCEvo\venue{CVPR'25} 
& Baseline implementation 
& \url{https://github.com/Beate-Suy-Zhang/DCEvo.git} 
& Not specified \\

Deno-IF\venue{NIPS'25} 
& Baseline implementation 
& \url{https://github.com/hanna-xu/Deno-IF.git} 
& Not specified \\

SAGE\venue{CVPR'25} 
& Baseline implementation 
& \url{https://github.com/RollingPlain/SAGE_IVIF.git} 
& MIT \\

UPFusion\venue{AAAI'26} 
& Baseline implementation 
& \url{https://github.com/ixilai/UP-Fusion.git} 
& Not specified \\

MSRS 
& Dataset 
& \url{https://github.com/Linfeng-Tang/MSRS.git} 
& Not specified \\

FMB 
& Dataset 
& \url{https://github.com/JinyuanLiu-CV/SegMiF.git} 
& Not specified \\

M3FD 
& Dataset 
& \url{https://github.com/JinyuanLiu-CV/TarDAL.git} 
& See TarDAL repository; dataset terms not separately specified \\

IVIF\_ZOO metrics 
& Evaluation code 
& \url{https://github.com/RollingPlain/IVIF_ZOO} 
& Not specified \\

SegFormer-B2 Cityscapes checkpoint 
& Downstream pretrained checkpoint 
& \url{https://huggingface.co/nvidia/segformer-b2-finetuned-cityscapes-1024-1024/tree/main} 
& Other \\
\bottomrule
\end{tabular}%
}
\end{table*}

\begin{table*}[t]
\centering
\caption{Complete image-level fusion results on MSRS. The best results are highlighted in \textbf{bold}, and the second-best results are \underline{underlined}.}
\label{tab:app_msrs_complete_1}
\setlength{\tabcolsep}{3pt}
\renewcommand{\arraystretch}{1.1}
\resizebox{\textwidth}{!}{%
\begin{tabular}{l|cccccccccc}
\toprule
Method 
& EN$\uparrow$ & AG$\uparrow$ & MI$\uparrow$ & SD$\uparrow$ 
& SCD$\uparrow$ & VIF$\uparrow$ & Qabf$\uparrow$ 
& SSIM$\uparrow$ & SF$\uparrow$ & CE$\downarrow$ \\
\midrule
SwinFusion\venue{JAS'22}
& 6.620 & 3.562 & 4.393 & 42.980 & \underline{1.686} & 1.032 & 0.647 & \underline{0.721} & 11.096 & 0.052562 \\

TarDAL\venue{CVPR'22}
& 6.211 & 2.926 & 2.669 & 32.781 & 1.375 & 0.716 & 0.416 & 0.540 & 9.547 & 0.065997 \\

CDDFuse\venue{CVPR'23}
& 6.699 & 3.749 & \underline{4.705} & 43.368 & 1.618 & 1.084 & 0.676 & 0.709 & 11.567 & 0.053406 \\

Diff-IF\venue{Inf. Fusion'24}
& 6.669 & 3.696 & \textbf{4.970} & 42.596 & 1.624 & \underline{1.097} & 0.685 & 0.711 & 11.460 & 0.054479 \\

EMMA\venue{CVPR'24}
& \textbf{6.715} & \underline{3.786} & 4.129 & \textbf{44.568} & 1.628 & 1.016 & 0.632 & 0.711 & \underline{11.574} & 0.052032 \\

DCEvo\venue{CVPR'25}
& 6.622 & 3.725 & 4.157 & 42.926 & 1.662 & 1.066 & \underline{0.688} & \underline{0.721} & 11.470 & \textbf{0.050816} \\

Deno-IF\venue{NIPS'25}
& 6.693 & 3.112 & 2.957 & 41.323 & \textbf{1.791} & 0.902 & 0.560 & 0.710 & 9.727 & 0.053525 \\

SAGE\venue{CVPR'25}
& 6.004 & 3.156 & 3.218 & 36.338 & 1.417 & 0.762 & 0.537 & 0.689 & 10.420 & 0.065907 \\

UPFusion\venue{AAAI'26}
& 6.608 & 3.562 & 3.618 & \underline{43.701} & 1.610 & 1.009 & 0.637 & \textbf{0.725} & 11.225 & \underline{0.051665} \\

\rowcolor{gray!20}
PCDFusion\venue{Ours}
& \underline{6.711} & \textbf{3.839} & 4.578 & 42.869 & 1.671 & \textbf{1.117} & \textbf{0.723} & 0.701 & \textbf{11.632} & 0.053841 \\
\bottomrule
\end{tabular}%
}
\end{table*}

\begin{table*}[t]
\centering
\caption{Complete image-level fusion results on FMB. The best results are highlighted in \textbf{bold}, and the second-best results are \underline{underlined}.}
\label{tab:app_fmb_complete_1}
\setlength{\tabcolsep}{3pt}
\renewcommand{\arraystretch}{1.1}
\resizebox{\textwidth}{!}{%
\begin{tabular}{l|cccccccccc}
\toprule
Method 
& EN$\uparrow$ & AG$\uparrow$ & MI$\uparrow$ & SD$\uparrow$ 
& SCD$\uparrow$ & VIF$\uparrow$ & Qabf$\uparrow$ 
& SSIM$\uparrow$ & SF$\uparrow$ & CE$\downarrow$ \\
\midrule
SwinFusion\venue{JAS'22}
& 6.683 & 4.052 & 4.270 & 35.352 & 1.560 & 1.032 & 0.665 & 0.743 & 13.515 & 0.000238 \\

TarDAL\venue{CVPR'22}
& \textbf{6.963} & 3.541 & 3.423 & \textbf{40.316} & 1.573 & 0.807 & 0.462 & 0.745 & 11.718 & 0.000369 \\

CDDFuse\venue{CVPR'23}
& 6.778 & 4.306 & 4.148 & 37.043 & 1.622 & 1.061 & 0.672 & 0.745 & \underline{14.577} & 0.000230 \\

Diff-IF\venue{Inf. Fusion'24}
& 6.632 & 4.047 & \textbf{4.985} & 34.191 & 1.368 & 1.047 & 0.660 & 0.735 & 13.869 & 0.000229 \\

EMMA\venue{CVPR'24}
& 6.775 & \textbf{4.684} & 3.947 & 36.805 & 1.504 & 1.031 & 0.645 & 0.735 & \textbf{15.006} & 0.000224 \\

DCEvo\venue{CVPR'25}
& 6.673 & 4.051 & \underline{4.512} & 34.749 & 1.503 & 1.058 & \underline{0.686} & 0.747 & 13.771 & 0.000218 \\

Deno-IF\venue{NIPS'25}
& 6.732 & 3.190 & 3.004 & 33.266 & \underline{1.654} & 0.830 & 0.568 & \textbf{0.767} & 11.243 & 0.000243 \\

SAGE\venue{CVPR'25}
& \underline{6.829} & 3.785 & 3.429 & 37.587 & \textbf{1.709} & 0.969 & 0.637 & \underline{0.752} & 12.644 & 0.000243 \\

UPFusion\venue{AAAI'26}
& 6.621 & 4.097 & 4.217 & 33.968 & 1.427 & \underline{1.081} & 0.684 & 0.747 & 13.892 & \underline{0.000212} \\

\rowcolor{gray!20}
PCDFusion\venue{Ours}
& 6.832 & \underline{4.334} & 4.411 & \underline{38.127} & 1.568 & \textbf{1.084} & \textbf{0.705} & 0.730 & 14.395 & \textbf{0.000186} \\
\bottomrule
\end{tabular}%
}
\end{table*}

\begin{table*}[t]
\centering
\caption{Complete image-level fusion results on M3FD. The best results are highlighted in \textbf{bold}, and the second-best results are \underline{underlined}.}
\label{tab:app_m3fd_complete_1}
\setlength{\tabcolsep}{3pt}
\renewcommand{\arraystretch}{1.1}
\resizebox{\textwidth}{!}{%
\begin{tabular}{l|cccccccccc}
\toprule
Method 
& EN$\uparrow$ & AG$\uparrow$ & MI$\uparrow$ & SD$\uparrow$ 
& SCD$\uparrow$ & VIF$\uparrow$ & Qabf$\uparrow$ 
& SSIM$\uparrow$ & SF$\uparrow$ & CE$\downarrow$ \\
\midrule
SwinFusion\venue{JAS'22}
& 6.808 & 4.606 & 4.020 & 35.776 & 1.560 & 0.898 & 0.607 & 0.717 & 13.672 & 0.031855 \\

TarDAL\venue{CVPR'22}
& \underline{6.985} & 4.265 & 3.130 & \textbf{38.785} & 1.523 & 0.708 & 0.427 & 0.708 & 12.672 & \textbf{0.029532} \\

CDDFuse\venue{CVPR'23}
& 6.907 & 4.861 & 3.774 & 37.206 & 1.646 & 0.917 & 0.610 & 0.718 & 14.753 & 0.031766 \\

Diff-IF\venue{Inf. Fusion'24}
& 6.791 & 4.567 & \textbf{4.852} & 34.927 & 1.322 & 0.894 & 0.589 & 0.701 & 14.115 & 0.031833 \\

EMMA\venue{CVPR'24}
& 6.927 & \textbf{5.335} & 3.753 & 38.253 & 1.494 & 0.897 & 0.590 & 0.703 & \textbf{15.235} & 0.031770 \\

DCEvo\venue{CVPR'25}
& 6.842 & 4.620 & \underline{4.298} & 35.582 & 1.491 & 0.917 & 0.633 & 0.720 & 14.109 & 0.031699 \\

Deno-IF\venue{NIPS'25}
& 6.789 & 3.713 & 2.701 & 32.569 & \underline{1.683} & 0.736 & 0.528 & \textbf{0.734} & 11.499 & \underline{0.031226} \\

SAGE\venue{CVPR'25}
& 6.850 & 4.249 & 3.115 & 36.465 & \textbf{1.694} & 0.856 & 0.583 & \underline{0.724} & 12.729 & 0.031960 \\

UPFusion\venue{AAAI'26}
& 6.819 & 4.697 & 4.023 & 35.611 & 1.394 & \textbf{0.954} & \underline{0.637} & 0.720 & 14.154 & 0.031829 \\

\rowcolor{gray!20}
PCDFusion\venue{Ours}
& \textbf{6.987} & \underline{5.037} & 3.962 & \underline{38.388} & 1.575 & \underline{0.949} & \textbf{0.668} & 0.699 & \underline{15.216} & 0.031670 \\
\bottomrule
\end{tabular}%
}
\end{table*}

\noindent\textbf{Baseline and Metric Evaluation.}
\label{app:baseline_metric}
For all compared IVIF methods, we use the official implementations and the publicly released original best checkpoints provided by the authors, without retraining or parameter tuning. 
Fusion results are generated with the default inference settings of each official codebase on the same test image pairs. 
We evaluate AG, SCD, VIF, Qabf, and Qcb using the public metric code from IVIF\_ZOO. 
For each dataset, each metric is computed on all test images and averaged over the dataset.

\noindent\textbf{Downstream Evaluation Protocol.}
\label{app:downstream_protocol}
For downstream evaluation, semantic and detection labels are used only to train task models, not to train PCDFusion. 
For FMB semantic segmentation, since FMB uses the same semantic categories as Cityscapes, we initialize SegFormerB2 with a Cityscapes-finetuned checkpoint. 
For M3FD object detection, we first train a YOLOv10s detector on the M3FD training split using both visible and infrared images for $100$ epochs, and use the resulting detector as the pretrained initialization.

To ensure a fair comparison, we generate fused training images for each compared method on the FMB and M3FD training sets. 
Starting from the same pretrained task model, we then fine-tune the segmentation or detection model on each method's fused training images for $50$ epochs. 
Finally, the fine-tuned task model is evaluated on the corresponding fused test images. 
All compared fusion methods use the same downstream training data, task labels, pretrained initialization, fine-tuning schedule, and evaluation protocol.

\section{Additional Quantitative Results}
\label{app:additional_quantitative}

\noindent\textbf{Complete Image-level Fusion Results.}
\label{app:complete_fusion_results}
\cref{tab:app_msrs_complete_1,tab:app_fmb_complete_1,tab:app_m3fd_complete_1} provide complete image-level metrics on MSRS, FMB, and M3FD. 
PCDFusion shows consistent advantages on fidelity and edge-preservation metrics. 
It achieves the best VIF and Qabf on MSRS and FMB, and obtains the best Qabf with the second-best VIF on M3FD. 
It also remains strong on gradient- and structure-related metrics, ranking first in AG and SF on MSRS, second in AG and SD on FMB, and second in AG, SD, and SF on M3FD. 
Although some competing methods lead in individual metrics such as MI, SCD, or SSIM, their advantages are usually dataset- or metric-specific. 
Overall, the complete results support that PCDFusion provides a stable fusion balance rather than optimizing a single quality indicator.

\begin{table*}[t]
\centering
\caption{Complete FMB semantic segmentation results. The zero-valued classes are omitted. The best results are highlighted in \textbf{bold}, and the second-best results are \underline{underlined}.}
\label{tab:app_fmb_seg_complete}
\setlength{\tabcolsep}{2pt}
\renewcommand{\arraystretch}{1.1}
\resizebox{\textwidth}{!}{%
\begin{tabular}{l|ccccccccccccc|c}
\toprule
Method 
& Road & Side. & Build. & Light & Sign & Veg. & Sky & Per. & Car & Truck & Bus & Motor. & Pole & mIoU$\uparrow$ \\
\midrule
SwinFusion\venue{JAS'22}
& 89.22 & 48.98 & 83.16 & 43.22 & 80.68 & 87.16 & 94.80 & 70.58 & 84.23 & \underline{48.77} & 57.57 & 36.28 & 50.23 & 58.33 \\

TarDAL\venue{CVPR'22}
& 89.50 & 50.05 & 83.24 & 42.30 & 81.20 & 88.21 & 95.00 & \textbf{73.31} & 81.85 & 35.82 & 26.36 & \underline{37.18} & 49.76 & 55.59 \\

CDDFuse\venue{CVPR'23}
& \textbf{90.18} & \underline{50.97} & 84.38 & 40.85 & 81.80 & \textbf{88.40} & 95.16 & 72.22 & 82.90 & 22.14 & 55.70 & 33.60 & \underline{51.56} & 56.66 \\

Diff-IF\venue{Inf. Fusion'24}
& 89.66 & 43.81 & 83.64 & 45.24 & \underline{82.04} & 87.02 & 94.70 & 70.69 & 83.57 & 45.22 & \textbf{75.99} & 32.26 & 50.82 & \underline{58.98} \\

EMMA\venue{CVPR'24}
& 89.61 & 49.32 & \textbf{84.57} & \textbf{49.42} & 81.24 & 88.15 & \textbf{95.29} & 70.75 & 82.58 & 42.99 & 59.41 & 32.61 & 51.07 & 58.47 \\

DCEvo\venue{CVPR'25}
& 89.72 & 48.01 & \underline{84.39} & \underline{46.98} & 81.01 & 88.14 & \underline{95.27} & 71.50 & \textbf{84.55} & 43.67 & 49.08 & 33.89 & 49.73 & 57.73 \\

Deno-IF\venue{NIPS'25}
& 88.89 & 47.39 & 84.05 & 40.84 & 80.57 & \underline{88.34} & 95.01 & \underline{72.73} & 82.25 & 40.13 & 34.40 & 29.23 & 47.86 & 55.45 \\

SAGE\venue{CVPR'25}
& \underline{89.97} & \textbf{51.65} & 83.45 & 42.90 & \textbf{82.37} & 88.08 & 94.90 & 71.45 & 82.28 & 41.08 & 31.51 & 33.66 & \textbf{51.61} & 56.33 \\

UPFusion\venue{AAAI'26}
& 88.83 & 43.86 & 84.23 & 19.68 & 77.97 & 87.64 & 95.00 & 69.73 & 81.92 & 39.27 & 58.29 & 34.85 & 49.21 & 55.36 \\

\rowcolor{gray!20}
PCDFusion\venue{Ours}
& 89.37 & 50.79 & 84.36 & 45.89 & 81.31 & 88.00 & 95.07 & 70.23 & \underline{84.44} & \textbf{51.54} & \underline{71.19} & \textbf{37.98} & 50.60 & \textbf{60.05} \\
\bottomrule
\end{tabular}%
}
\end{table*}

\begin{table*}[t]
\centering
\caption{Complete M3FD object detection results. The best results are highlighted in \textbf{bold}, and the second-best results are \underline{underlined}.}
\label{tab:app_m3fd_det_complete}
\setlength{\tabcolsep}{2pt}
\renewcommand{\arraystretch}{1.1}
\resizebox{\textwidth}{!}{%
\begin{tabular}{l|ccccccc|ccccccc}
\toprule
\multirow{2}{*}{Method}
& \multicolumn{7}{c|}{mAP$_{50}$}
& \multicolumn{7}{c}{mAP$_{50:95}$} \\
\cmidrule(lr){2-8} \cmidrule(lr){9-15}
& Per. & Car & Bus & Motor. & Lamp & Truck & mAP
& Per. & Car & Bus & Motor. & Lamp & Truck & mAP \\
\midrule
SwinFusion\venue{JAS'22}
& 82.28 & 91.25 & 89.10 & 66.22 & 79.83 & 78.75 & 81.24
& 48.90 & 68.13 & 71.31 & 40.82 & 42.03 & 58.00 & 54.86 \\

TarDAL\venue{CVPR'22}
& 81.94 & 90.88 & 87.43 & 59.35 & 78.67 & 76.25 & 79.09
& 48.80 & 66.51 & 69.55 & 34.00 & 38.87 & 56.38 & 52.35 \\

CDDFuse\venue{CVPR'23}
& 81.93 & 91.57 & \underline{90.26} & 68.36 & 81.59 & 74.18 & 81.31
& 48.79 & 68.44 & \underline{71.80} & 38.00 & 43.44 & 54.39 & 54.15 \\

Diff-IF\venue{Inf. Fusion'24}
& 82.81 & 91.38 & 88.46 & 66.32 & 77.91 & \underline{79.14} & 81.00
& 49.77 & 67.43 & 70.08 & 38.47 & 39.82 & \underline{58.68} & 54.04 \\

EMMA\venue{CVPR'24}
& 81.70 & 91.11 & \textbf{90.46} & 63.35 & \underline{81.59} & 77.36 & 80.93
& 48.35 & 68.03 & 71.02 & 38.26 & 43.70 & 57.20 & 54.43 \\

DCEvo\venue{CVPR'25}
& \textbf{83.59} & 91.52 & 88.96 & \textbf{71.23} & 78.56 & \textbf{80.10} & \textbf{82.33}
& \textbf{50.05} & 68.25 & 71.60 & \underline{41.20} & 43.86 & \textbf{59.78} & \underline{55.79} \\

Deno-IF\venue{NIPS'25}
& \underline{83.34} & \underline{91.88} & 88.38 & \underline{69.87} & 77.82 & 78.98 & \underline{81.71}
& \underline{50.03} & \underline{68.68} & 70.37 & 41.06 & 43.16 & 57.61 & 55.15 \\

SAGE\venue{CVPR'25}
& 82.34 & 91.30 & 86.46 & 65.82 & 78.10 & 75.46 & 79.91
& 49.59 & 68.42 & 68.91 & 39.15 & 43.22 & 56.39 & 54.28 \\

UPFusion\venue{AAAI'26}
& 82.09 & 91.82 & 87.90 & 65.47 & \textbf{82.29} & 76.73 & 81.05
& 48.72 & 68.52 & 71.28 & 38.57 & \textbf{45.94} & 58.04 & 55.18 \\

\rowcolor{gray!20}
PCDFusion\venue{Ours}
& 82.30 & \textbf{91.93} & 90.04 & 69.53 & 79.38 & 76.99 & 81.69
& 48.95 & \textbf{69.00} & \textbf{72.60} & \textbf{42.89} & \underline{44.53} & 57.93 & \textbf{55.98} \\
\bottomrule
\end{tabular}%
}
\end{table*}

\noindent\textbf{Complete downstream results.}
\cref{tab:app_fmb_seg_complete,tab:app_m3fd_det_complete} report complete downstream results on FMB semantic segmentation and M3FD object detection. 
For segmentation, zero-valued classes are omitted from the table for readability, while the reported mIoU follows the same evaluation setting as the main paper. 
PCDFusion achieves the best overall mIoU and performs strongly on object-related and thin-structure categories, including Car, Truck, Bus, and Motorcycle. 
For detection, PCDFusion obtains the best mAP$_{50:95}$ and the second-best mAP$_{50}$, with clear gains on Car, Bus, and Motorcycle under the stricter mAP$_{50:95}$ metric. 
These results indicate that the fused images produced by PCDFusion support both semantic structure parsing and object localization, rather than improving only image-level fusion scores.

\noindent\textbf{Local structure recovery and appearance preservation.}
The first MSRS example in \cref{fig:visual_msrs_fmb} contains a locally saturated building whose contour remains visible in the infrared observation. 
To separate useful structure recovery from indiscriminate infrared transfer, we measure infrared-supported edge retention on over-exposed visible pixels with strong infrared gradients and reliable-sky preservation on sky pixels whose fused luminance remains within $0.05$ of the visible input. 
As shown in \cref{tab:app_roi_analysis}, PCDFusion achieves the highest edge retention, $66.3\%$, while preserving $99.5\%$ of reliable-sky pixels. 
This combination indicates that the recovered contour is localized to the failed visible region rather than produced by globally copying infrared edges.

\begin{table*}[t]
\centering
\caption{Local structure recovery and appearance preservation in the first MSRS visualization. The best result is highlighted in \textbf{bold}, and the second-best result is \underline{underlined}.}
\label{tab:app_roi_analysis}
\setlength{\tabcolsep}{2.5pt}
\renewcommand{\arraystretch}{1.1}
\resizebox{\textwidth}{!}{%
\begin{tabular}{l|cccccccccc}
\toprule
Metric
& SwinFusion & TarDAL & CDDFuse & Diff-IF & EMMA
& DCEvo & Deno-IF & SAGE & UPFusion & PCDFusion \\
\midrule
Over-exposed IR-edge retention (\%)$\uparrow$
& 16.8 & 27.0 & 51.3 & 47.4 & 24.0
& \underline{62.9} & 49.0 & 25.7 & 17.3 & \textbf{66.3} \\
Reliable-sky preservation (\%)$\uparrow$
& \textbf{100.0} & 0.1 & 99.5 & 99.6 & 97.5
& 99.5 & 7.3 & 0.0 & \underline{99.9} & 99.5 \\
\bottomrule
\end{tabular}%
}
\end{table*}

\noindent\textbf{Single-modality-to-fused transfer.}
We additionally evaluate zero-shot compatibility with task models trained on individual modalities. 
Visible-only and infrared-only task models are trained separately and then evaluated directly on each method's fused test images; the reported mean is the arithmetic mean of the two scores. 
All methods use the same data splits, task architectures, and evaluation code, and each entry is averaged over three task-model seeds. 
\Cref{tab:app_single_modality_transfer} shows that PCDFusion ranks first on both aggregate means, reaching $48.77\%$ mIoU on FMB and $19.80\%$ mAP$_{50:95}$ on M3FD. 
The result complements the task-adapted protocol in the main paper by showing that the fused images also transfer well to models that have not been trained on fused data.

\begin{table*}[t]
\centering
\caption{Single-modality-to-fused transfer results. VIS-trained and IR-trained task models are evaluated directly on fused test images; Mean averages the two scores. The best result is highlighted in \textbf{bold}, and the second-best result is \underline{underlined}.}
\label{tab:app_single_modality_transfer}
\setlength{\tabcolsep}{5pt}
\renewcommand{\arraystretch}{1.1}
\begin{tabular}{l|ccc|ccc}
\toprule
\multirow{2}{*}{Fusion method}
& \multicolumn{3}{c|}{FMB mIoU (\%)}
& \multicolumn{3}{c}{M3FD mAP$_{50:95}$ (\%)} \\
\cmidrule(lr){2-4}\cmidrule(lr){5-7}
& VIS-trained & IR-trained & Mean
& VIS-trained & IR-trained & Mean \\
\midrule
SwinFusion & 54.18 & 40.39 & 47.28 & 25.96 & 8.46 & 17.21 \\
TarDAL     & 51.76 & \textbf{42.46} & 47.11 & 24.48 & 10.04 & 17.26 \\
CDDFuse    & \underline{56.09} & 40.79 & \underline{48.44} & 26.54 & 8.35 & 17.44 \\
Diff-IF    & 53.61 & 40.01 & 46.81 & 23.30 & \underline{10.34} & 16.82 \\
EMMA       & 55.65 & 39.59 & 47.62 & 30.01 & 5.93 & 17.97 \\
DCEvo      & 54.96 & 40.84 & 47.90 & 26.12 & 9.32 & 17.72 \\
Deno-IF    & 54.28 & 41.23 & 47.75 & 26.57 & \textbf{10.91} & 18.74 \\
SAGE       & 54.76 & \underline{41.45} & 48.10 & 26.99 & 9.89 & 18.44 \\
UPFusion   & 55.96 & 40.13 & 48.04 & \underline{32.28} & 6.71 & \underline{19.50} \\
\rowcolor{gray!20}
PCDFusion  & \textbf{56.36} & 41.18 & \textbf{48.77} & \textbf{32.30} & 7.31 & \textbf{19.80} \\
\bottomrule
\end{tabular}
\end{table*}

\noindent\textbf{Comparison with feature-level RGB-IR detection.}
Pixel-level fusion produces a reusable image for visualization and conventional single-image task models, whereas feature-level detectors consume paired visible and infrared inputs through a task-specific architecture. 
We compare these two paradigms on the same $300$-image M3FD test set using the same evaluation code and closely matched model capacity. 
As reported in \cref{tab:app_feature_level_detection}, PCDFusion with YOLOv10x reaches $61.05\%$ mAP$_{50:95}$, only $0.22$ points below M2D-LIF and $1.67$ points above LoME. 
Thus, PCDFusion approaches the strongest feature-level result while retaining a viewable fused output and single-image inference for downstream deployment.

\begin{table*}[t]
\centering
\caption{Capacity-matched comparison with feature-level RGB-IR detectors on M3FD. Parameter counts include the complete training architectures, and GFLOPs are measured for deployed $640\times640$ inference.}
\label{tab:app_feature_level_detection}
\setlength{\tabcolsep}{5pt}
\renewcommand{\arraystretch}{1.15}
\resizebox{\textwidth}{!}{%
\begin{tabular}{l|c|c|c|ccc}
\toprule
Method & Fusion level & Test-time inputs & Training protocol
& Params (M)$\downarrow$ & GFLOPs$\downarrow$ & mAP$_{50:95}$ (\%)$\uparrow$ \\
\midrule
PCDFusion + YOLOv10x
& Pixel/image & One fused image & 300 epochs
& 36.27 & 508.62 & 61.05 \\
M2D-LIF
& Feature & Paired VIS+IR & 200-epoch teacher + 100-epoch student
& 36.51 & 122.42 & \textbf{61.26} \\
LoME (width $0.90$)
& Feature & Paired VIS+IR & 300 epochs
& 36.18 & 49.05 & 59.38 \\
\bottomrule
\end{tabular}%
}
\end{table*}

\section{Additional Qualitative Results}
\label{app:additional_qualitative}
\begin{figure*}[t]
    \centering
    \includegraphics[width=\textwidth]{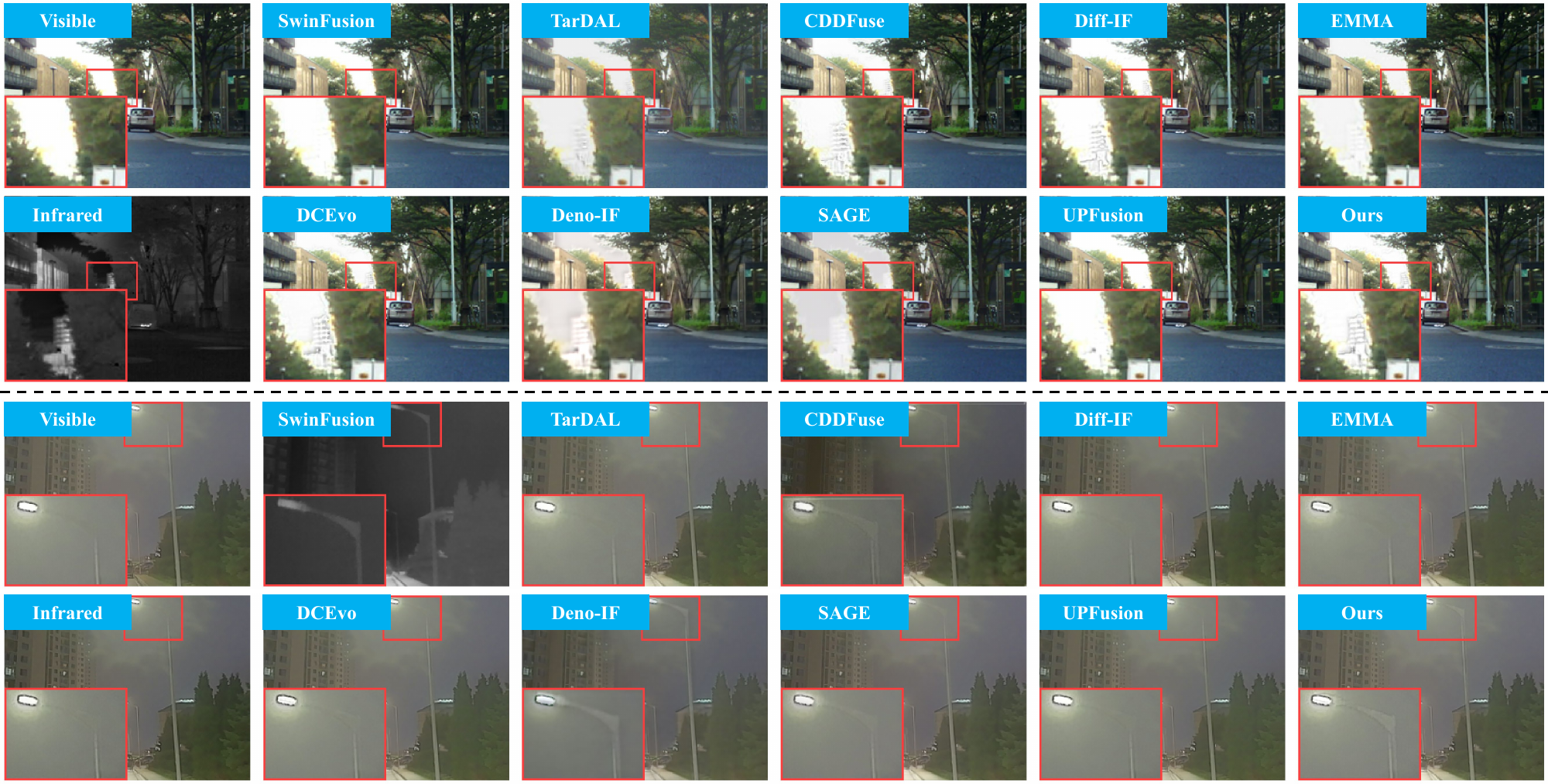}
    \caption{Additional visual comparisons on MSRS and FMB.}
    \label{fig:app_visual_results}
\end{figure*}

\noindent\textbf{More Image-level fusion visualizations.} 
\cref{fig:app_visual_results} provides additional qualitative comparisons on challenging scenes. 
In the first scene, the visible image contains an over-exposed building region, while the infrared image preserves its coarse structure. 
Most competing methods either keep the region saturated or recover only weak contours. 
PCDFusion better restores the building structure while maintaining the surrounding visible appearance. 
In the second scene, the infrared image clearly highlights the street lamp under low-contrast visible observation. 
PCDFusion effectively incorporates the lamp response without spreading excessive thermal intensity to the background.

\begin{figure*}[t]
    \centering
    \includegraphics[width=\textwidth]{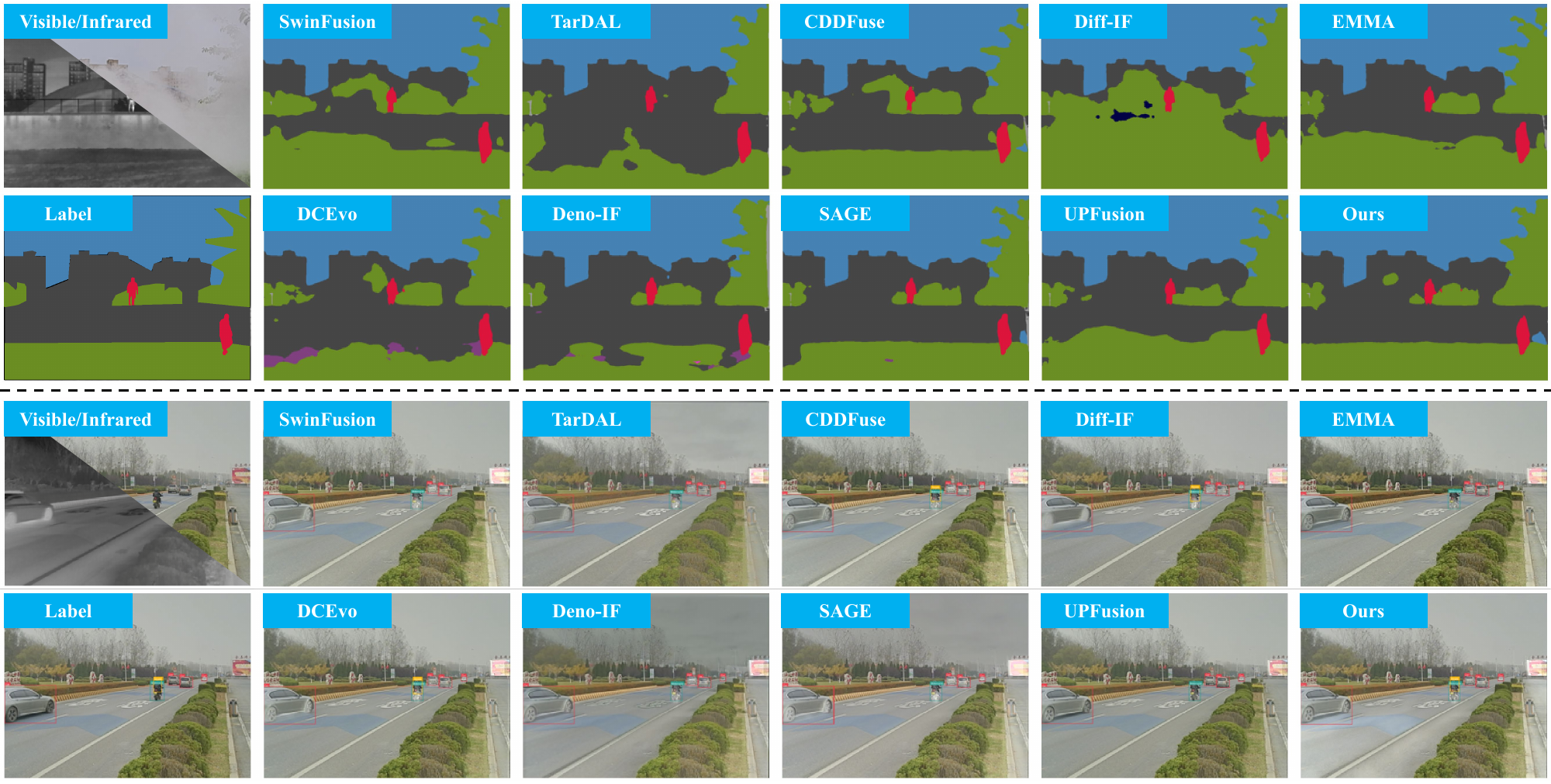}
    \caption{Additional FMB semantic segmentation and M3FD object detection visualizations.}
    \label{fig:app_visual_downs}
\end{figure*}

\noindent\textbf{More downstream visualizations.} 
\cref{fig:app_visual_downs} shows additional downstream results. 
For semantic segmentation, PCDFusion produces cleaner masks and almost completely segments the vegetation and pedestrians in the scene. 
For object detection, PCDFusion accurately detects the motorcycle and the rider on it, while avoiding obvious background interference. 
These results further indicate that the fused images preserve task-relevant structures and targets.

\section{Additional Discussions}
\label{app:additional_ablation}

\noindent\textbf{Additional Prior Variants.}
\label{app:prior_variants}
The main paper reports the ablation of posterior estimation forms and context prior estimates. 
Here, we further provide the ablation of proposal and detail prior variants. 
These variants test whether the stage-specific physical priors can be replaced by simpler estimates. 
We denote the normalized visible gradient by $\mathbf{S}_{v}=\operatorname{Norm}(|\nabla\mathbf{Y}_{v}|)$, the normalized infrared edge response by $\mathbf{S}_{er}=\operatorname{Norm}(|\nabla\mathbf{Y}_{r}|)$, and the combined infrared structure by $\mathbf{S}_{r}=\max(\mathbf{S}_{er},\mathbf{S}_{br})$, where $\mathbf{S}_{br}$ is the band-limited infrared response.
\begin{table*}[t]
\centering
\caption{Additional ablation of proposal and detail prior variants. The best results are highlighted in \textbf{bold}, and the second-best results are \underline{underlined}.}
\label{tab:app_prior_variants}
\setlength{\tabcolsep}{4pt}
\renewcommand{\arraystretch}{1.1}
\resizebox{\textwidth}{!}{%
\begin{tabular}{l|l|ccccc|cc}
\toprule
Stage & Variant 
& AG$\uparrow$ & SCD$\uparrow$ & VIF$\uparrow$ & Qabf$\uparrow$ & Qcb$\uparrow$ 
& mIoU$\uparrow$ & mAP$\uparrow$ \\
\midrule
Proposal & Uniform: $0.5$
& \textbf{3.863} & \textbf{1.672} & \underline{1.110} & \textbf{0.723} & \underline{0.591} & 58.88 & \underline{55.86} \\

Proposal & Hard IR: $\mathbbm{1}[\operatorname{Norm}(\mathbf{Y}_{r})\geq0.5]$
& \underline{3.861} & \underline{1.671} & 1.109 & 0.722 & 0.589 & \underline{58.95} & 55.72 \\

\rowcolor{gray!20}
Proposal & Soft IR (Full): $\sigma(\tau_p(\operatorname{Norm}(\mathbf{Y}_{r})-0.5))$
& 3.839 & \underline{1.671} & \textbf{1.117} & \textbf{0.723} & \textbf{0.593} & \textbf{60.05} & \textbf{55.98} \\

\midrule
Detail & Edge competition: $\mathbf{S}_{er}/(\mathbf{S}_{er}+\mathbf{S}_{v}+\epsilon)$
& \underline{3.855} & 1.674 & 1.104 & 0.721 & 0.585 & \underline{58.95} & 55.83 \\

Detail & IR structure only: $\mathbf{S}_{r}$
& 3.848 & \underline{1.677} & \underline{1.110} & \underline{0.722} & \underline{0.587} & 58.84 & 55.70 \\

Detail & IR intensity: $\operatorname{Norm}(\mathbf{Y}_{r})$
& \textbf{3.859} & \textbf{1.678} & 1.105 & 0.721 & 0.586 & 58.90 & \underline{55.95} \\

\rowcolor{gray!20}
Detail & Structure competition (Full): $\mathbf{S}_{r}/(\mathbf{S}_{r}+\mathbf{S}_{v}+\epsilon)$
& 3.839 & 1.671 & \textbf{1.117} & \textbf{0.723} & \textbf{0.593} & \textbf{60.05} & \textbf{55.98} \\
\bottomrule
\end{tabular}%
}
\end{table*}
For the proposal stage, constant or hard priors can slightly increase AG or SCD, but they reduce VIF, Qcb, mIoU, and mAP compared with the full model. 
This suggests that the soft thermal-response prior is more suitable for preserving possible thermal responses without over-constraining later rendering. 
For the detail stage, edge-, structure-, or intensity-based variants improve some difference-related scores, but the full model achieves stronger fidelity and downstream performance. 
This supports the use of structure-competitive detail prior together with posterior calibration, rather than directly rendering infrared edges or intensities.

\noindent\textbf{Fixed-checkpoint parameter sensitivity.}
To isolate inference-time sensitivity from retraining variation, we keep the Full checkpoint fixed and vary one scalar at a time. 
\Cref{tab:app_parameter_sensitivity} reports the absolute values together with their changes from the default. 
Across proposal temperatures from $2$ to $10$, FMB mIoU and M3FD mAP vary by only $0.12$ and $0.11$ points, respectively. 
Across detail scales from $0.05$ to $0.15$, the corresponding spans are $0.22$ and $0.29$ points. 
The stable downstream performance over these broad ranges indicates that the gains do not depend on a narrowly tuned inference scalar.

\begin{table*}[t]
\centering
\caption{Fixed-checkpoint parameter sensitivity. Parenthesized values show changes from the default setting.}
\label{tab:app_parameter_sensitivity}
\setlength{\tabcolsep}{4pt}
\renewcommand{\arraystretch}{1.12}
\resizebox{\textwidth}{!}{%
\begin{tabular}{l|c|ccccc}
\toprule
Parameter & Value & AG & VIF & Qabf & mIoU & mAP \\
\midrule
\multirow{5}{*}{Proposal temperature}
& $2$
& 3.813 {\color{gray}\scriptsize($-0.026$)}
& 1.115 {\color{gray}\scriptsize($-0.002$)}
& 0.720 {\color{gray}\scriptsize($-0.003$)}
& 60.11 {\color{gray}\scriptsize($+0.06$)}
& 55.94 {\color{gray}\scriptsize($-0.05$)} \\
& $4$
& 3.832 {\color{gray}\scriptsize($-0.007$)}
& 1.119 {\color{gray}\scriptsize($+0.002$)}
& 0.723 {\color{gray}\scriptsize($0.000$)}
& 60.10 {\color{gray}\scriptsize($+0.05$)}
& 55.88 {\color{gray}\scriptsize($-0.11$)} \\
& $6$ (default)
& 3.839 {\color{gray}\scriptsize($0.000$)}
& 1.117 {\color{gray}\scriptsize($0.000$)}
& 0.723 {\color{gray}\scriptsize($0.000$)}
& 60.05 {\color{gray}\scriptsize($0.00$)}
& 55.98 {\color{gray}\scriptsize($0.00$)} \\
& $8$
& 3.830 {\color{gray}\scriptsize($-0.009$)}
& 1.104 {\color{gray}\scriptsize($-0.013$)}
& 0.720 {\color{gray}\scriptsize($-0.002$)}
& 60.03 {\color{gray}\scriptsize($-0.02$)}
& 55.90 {\color{gray}\scriptsize($-0.08$)} \\
& $10$
& 3.794 {\color{gray}\scriptsize($-0.045$)}
& 1.080 {\color{gray}\scriptsize($-0.037$)}
& 0.714 {\color{gray}\scriptsize($-0.009$)}
& 59.99 {\color{gray}\scriptsize($-0.06$)}
& 55.95 {\color{gray}\scriptsize($-0.04$)} \\
\midrule
\multirow{5}{*}{Detail scale}
& $0.050$
& 3.729 {\color{gray}\scriptsize($-0.110$)}
& 1.109 {\color{gray}\scriptsize($-0.008$)}
& 0.714 {\color{gray}\scriptsize($-0.008$)}
& 59.90 {\color{gray}\scriptsize($-0.15$)}
& 55.88 {\color{gray}\scriptsize($-0.10$)} \\
& $0.075$
& 3.782 {\color{gray}\scriptsize($-0.057$)}
& 1.114 {\color{gray}\scriptsize($-0.003$)}
& 0.720 {\color{gray}\scriptsize($-0.003$)}
& 60.00 {\color{gray}\scriptsize($-0.05$)}
& 55.84 {\color{gray}\scriptsize($-0.14$)} \\
& $0.100$ (default)
& 3.839 {\color{gray}\scriptsize($0.000$)}
& 1.117 {\color{gray}\scriptsize($0.000$)}
& 0.723 {\color{gray}\scriptsize($0.000$)}
& 60.05 {\color{gray}\scriptsize($0.00$)}
& 55.98 {\color{gray}\scriptsize($0.00$)} \\
& $0.125$
& 3.898 {\color{gray}\scriptsize($+0.060$)}
& 1.118 {\color{gray}\scriptsize($+0.001$)}
& 0.722 {\color{gray}\scriptsize($-0.001$)}
& 60.11 {\color{gray}\scriptsize($+0.06$)}
& 56.00 {\color{gray}\scriptsize($+0.02$)} \\
& $0.150$
& 3.960 {\color{gray}\scriptsize($+0.122$)}
& 1.117 {\color{gray}\scriptsize($0.000$)}
& 0.720 {\color{gray}\scriptsize($-0.003$)}
& 60.12 {\color{gray}\scriptsize($+0.07$)}
& 55.72 {\color{gray}\scriptsize($-0.27$)} \\
\bottomrule
\end{tabular}%
}
\end{table*}

\noindent\textbf{Stage and objective trade-offs.}
The metric reversals in the stage ablation follow from the distinct image properties modified by the three renderers. 
Proposal rendering maintains high-recall thermal candidates; removing it discards target evidence before later selection and lowers mIoU and mAP by $1.61$ and $0.80$ points. 
Context rendering repairs exposure-failed visible regions and suppresses unsupported variation. 
Without it, SCD increases from $1.671$ to $1.716$ because more cross-modal contrast remains, but mIoU and mAP decrease by $1.19$ and $1.10$ points. 
Detail rendering injects a signed band-limited residual under structure competition, improving VIF, Qcb, and mIoU while allowing a small decrease in AG. 
The $0.15$-point mAP advantage without this stage reflects that detector confidence is not optimized directly and need not vary monotonically with all fusion-quality measures. 
The Full model therefore provides the strongest overall balance rather than maximizing every scalar independently.

The objective ablation exhibits a related hierarchy. 
$\mathcal{L}_{d}$ supervises the final fused luminance densely and directly, whereas $\mathcal{L}_{p}$, $\mathcal{L}_{c}$, and the keep term constrain intermediate proposal and context decisions. 
When one intermediate constraint is removed, later renderers and $\mathcal{L}_{d}$ can compensate for part of the missing guidance, producing smaller changes in aggregate downstream scores. 
Nevertheless, these terms prevent different failure modes: $\mathcal{L}_{p}$ preserves infrared-salient candidates, $\mathcal{L}_{c}$ anchors context updates to visible failure with infrared support, and the keep term suppresses unnecessary updates in reliable visible regions. 
Their joint use preserves the intended proposal-context-detail behavior across fusion and downstream criteria.

\noindent\textbf{Posterior-action interventions.}
We directly test whether the spatial values of the posterior maps contribute to local rendering quality. 
With the checkpoint and inputs fixed, one learned posterior-action map at a time is replaced by its prior, a spatially shuffled map, an inverted map, or a zero map. 
As shown in \cref{tab:app_posterior_intervention}, the learned posterior provides the largest local task gain at every stage and the best final FMB mIoU in every intervention group. 
Shuffling the proposal or context map reduces mIoU from $60.05$ to $50.21$ or $50.16$, respectively, demonstrating that the observation-conditioned spatial assignment, rather than only the mean weight magnitude, is important for the fused result.

\begin{table*}[t]
\centering
\caption{Posterior-action interventions and local utility. Each intervention replaces one stage map while keeping the checkpoint and inputs fixed. Local gains and FMB mIoU are reported in percentage points.}
\label{tab:app_posterior_intervention}
\setlength{\tabcolsep}{8pt}
\renewcommand{\arraystretch}{1.1}
\begin{tabular}{l|l|cc}
\toprule
Stage & Posterior-action map used for rendering & Local task gain$\uparrow$ & FMB mIoU$\uparrow$ \\
\midrule
\multirow{5}{*}{Proposal}
& Learned posterior & \textbf{0.071} & \textbf{60.05} \\
& Prior only & 0.069 & 59.86 \\
& Spatially shuffled & $-0.491$ & 50.21 \\
& Inverted & 0.003 & 58.68 \\
& Zero map & 0.000 & 58.58 \\
\midrule
\multirow{5}{*}{Context}
& Learned posterior & \textbf{0.013} & \textbf{60.05} \\
& Prior only & 0.003 & 59.36 \\
& Spatially shuffled & $-0.519$ & 50.16 \\
& Inverted & $-0.020$ & 57.66 \\
& Zero map & 0.000 & 58.68 \\
\midrule
\multirow{5}{*}{Detail}
& Learned posterior & \textbf{0.0029} & \textbf{60.05} \\
& Prior only & 0.0014 & 59.97 \\
& Spatially shuffled & 0.0018 & 59.73 \\
& Inverted & $-0.0004$ & 59.42 \\
& Zero map & 0.0000 & 59.54 \\
\bottomrule
\end{tabular}
\end{table*}

\begin{table*}[t]
\centering
\caption{Computational cost comparison on $512\times512$ inputs. The best results are highlighted in \textbf{bold}, and the second-best results are \underline{underlined}.}
\label{tab:app_efficiency}
\setlength{\tabcolsep}{3pt}
\renewcommand{\arraystretch}{1.1}
\resizebox{\textwidth}{!}{%
\begin{tabular}{l|cccccccccc}
\toprule
Metric
& SwinFusion\venue{JAS'22}
& TarDAL\venue{CVPR'22}
& CDDFuse\venue{CVPR'23}
& Diff-IF\venue{Inf. Fusion'24}
& EMMA\venue{CVPR'24}
& DCEvo\venue{CVPR'25}
& Deno-IF\venue{NIPS'25}
& SAGE\venue{CVPR'25}
& UPFusion\venue{AAAI'26}
& PCDFusion\venue{Ours} \\
\midrule
Params (M)$\downarrow$
& 0.97 & \underline{0.30} & 1.19 & 23.74 & 1.52 & 2.00 & 1.43 & \textbf{0.14} & 154.82 & 4.60 \\

GFLOPs$\downarrow$
& 604.42 & 155.07 & 945.04 & 6101.12 & \underline{68.46} & 1561.87 & 221.15 & \textbf{34.11} & 2676.36 & 235.34 \\

Time (ms)$\downarrow$
& 997.82 & \underline{5.04} & 126.70 & 780.17 & 13.17 & 178.92 & 112.68 & \textbf{2.12} & 584.92 & 92.51 \\

FPS$\uparrow$
& 1.00 & \underline{198.55} & 7.89 & 1.28 & 75.92 & 5.59 & 8.87 & \textbf{471.34} & 1.71 & 10.81 \\
\bottomrule
\end{tabular}%
}
\end{table*}

\noindent\textbf{Computational cost.}
\Cref{tab:app_efficiency} reports the model size, GFLOPs, inference time, and FPS under the same $512\times512$ input setting. 
PCDFusion has $4.60$M parameters and requires $235.34$ GFLOPs, which is higher than several lightweight baselines but much lower than large generative or heavy interaction-based methods such as Diff-IF and UPFusion. 
Its inference speed is also substantially faster than SwinFusion, Diff-IF, and UPFusion. 
This indicates that PCDFusion achieves the reported fusion and downstream gains with moderate computational cost, although it is not designed as the fastest lightweight model.

\begin{table*}[t]
\centering
\caption{Calibration of exposure-failure estimates on MSRS and FMB. The best results are highlighted in \textbf{bold}, and the second-best results are \underline{underlined}.}
\label{tab:app_prior_calibration}
\setlength{\tabcolsep}{3pt}
\renewcommand{\arraystretch}{1.1}
\resizebox{\textwidth}{!}{%
\begin{tabular}{l|l|ccc|ccc}
\toprule
\multirow{2}{*}{Dataset}
& \multirow{2}{*}{Estimate}
& \multicolumn{3}{c|}{Pixel failure}
& \multicolumn{3}{c}{Class-normalized mFailure} \\
\cmidrule(lr){3-5} \cmidrule(lr){6-8}
& 
& Pearson$\uparrow$ & Spearman$\uparrow$ & MSE$\downarrow$
& Pearson$\uparrow$ & Spearman$\uparrow$ & MSE$\downarrow$ \\
\midrule
\multirow{5}{*}{MSRS}
& Darkness-only: $1-\mathbf{Y}_{v}$
& \textbf{0.762} & \textbf{0.982} & 0.1372
& -0.111 & -0.035 & 0.2231 \\

& Brightness-only: $\mathbf{Y}_{v}$
& -0.762 & -0.982 & 0.3381
& 0.111 & 0.035 & 0.1936 \\

& Distance-to-mid: $\left|2\mathbf{Y}_{v}-1\right|$
& 0.418 & 0.087 & 0.1825
& 0.815 & \textbf{0.928} & 0.0998 \\

& Ours prior: $\left(2\mathbf{Y}_{v}-1\right)^2$
& 0.463 & 0.087 & \underline{0.1014}
& \underline{0.879} & \textbf{0.928} & \textbf{0.0314} \\

& Ours posterior: $\mathbf{q}_{c}$
& \underline{0.646} & \underline{0.147} & \textbf{0.0552}
& \textbf{0.890} & \underline{0.904} & \underline{0.0658} \\
\midrule
\multirow{5}{*}{FMB}
& Darkness-only: $1-\mathbf{Y}_{v}$
& 0.524 & 0.432 & 0.1286
& \textbf{0.563} & \textbf{0.555} & \textbf{0.0691} \\

& Brightness-only: $\mathbf{Y}_{v}$
& -0.524 & -0.432 & 0.2849
& -0.563 & -0.555 & 0.2393 \\

& Distance-to-mid: $\left|2\mathbf{Y}_{v}-1\right|$
& 0.530 & \underline{0.490} & 0.1261
& 0.369 & 0.405 & \underline{0.0983} \\

& Ours prior: $\left(2\mathbf{Y}_{v}-1\right)^2$
& \underline{0.565} & \underline{0.490} & \underline{0.0722}
& 0.445 & 0.405 & 0.1358 \\

& Ours posterior: $\mathbf{q}_{c}$
& \textbf{0.661} & \textbf{0.737} & \textbf{0.0412}
& \underline{0.507} & \underline{0.435} & 0.1733 \\
\bottomrule
\end{tabular}%
}
\end{table*}

\begin{figure*}[t]
    \centering
    \begin{minipage}[t]{0.48\textwidth}
        \centering
        \includegraphics[width=\linewidth]{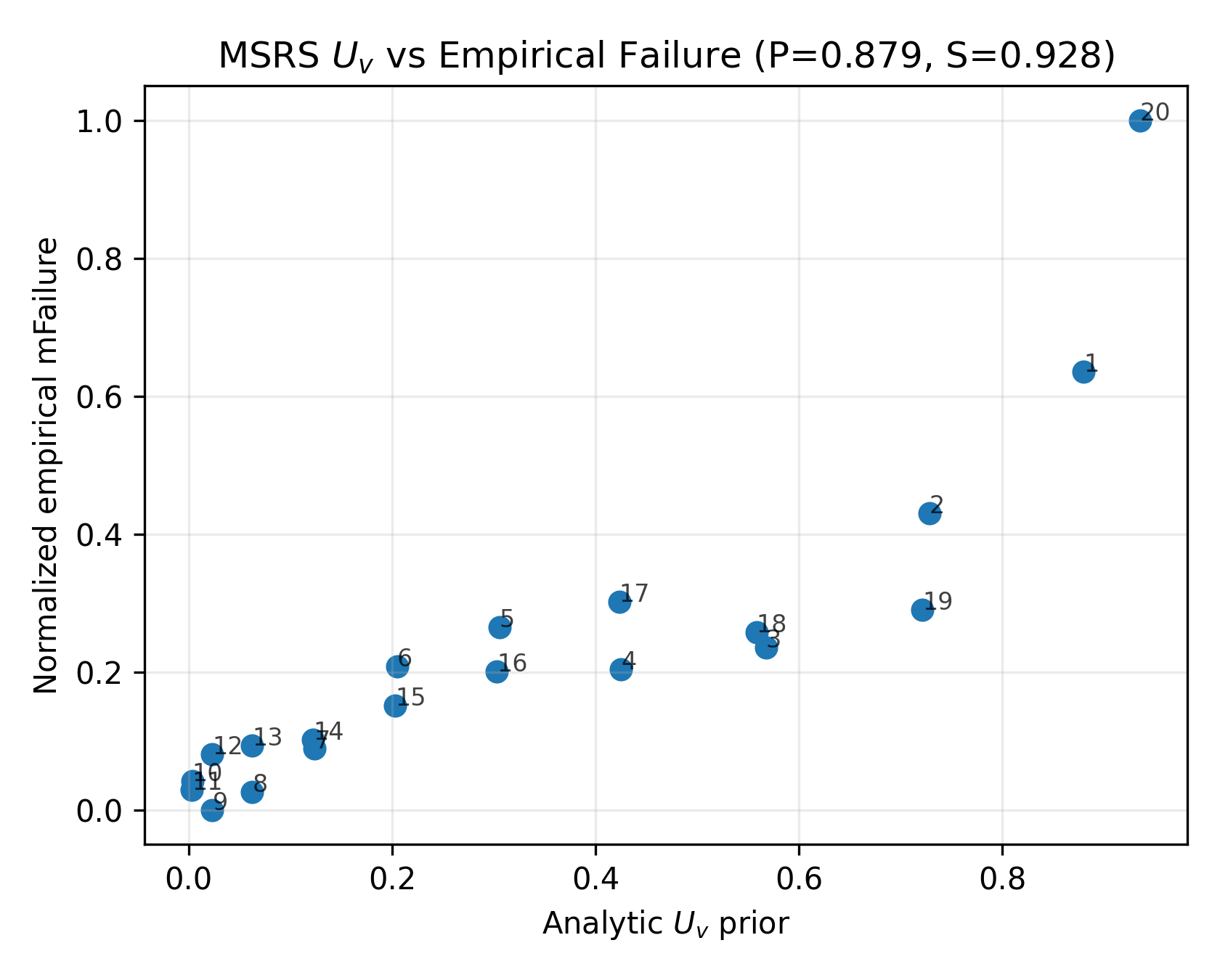}
    \end{minipage}
    \hfill
    \begin{minipage}[t]{0.48\textwidth}
        \centering
        \includegraphics[width=\linewidth]{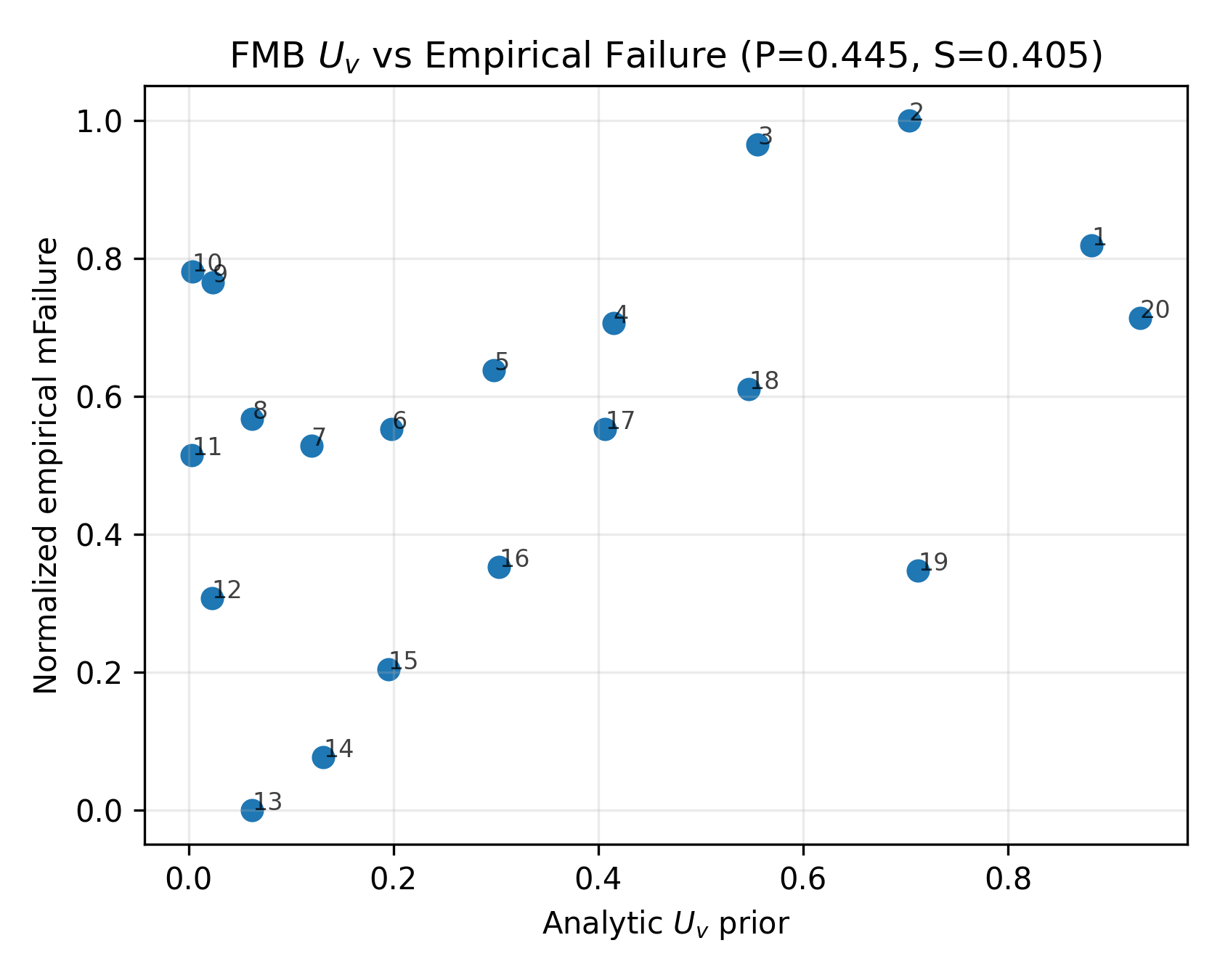}
    \end{minipage}
    \caption{
    Relation between the analytic exposure-failure prior $U_v$ and empirical class-normalized visible failure on MSRS and FMB.
    Each point denotes a luminance bin, and Pearson/Spearman correlations are reported in the subplot titles.
    }
    \label{fig:app_uv_empirical_failure}
\end{figure*}

\noindent\textbf{Exposure-Failure Prior Calibration.}
\label{app:prior_calibration}
We further examine whether the analytic exposure-failure prior agrees with empirical visible failures. 
This analysis is only used for post-hoc validation and is not used to train PCDFusion. 
We use a semantic segmentation model fine-tuned on MSRS to estimate how often visible observations fail under different luminance levels.

For each visible image, let $\mathbf{Y}_{v}$ be the visible luminance, $\hat{\mathbf{M}}_{v}$ be the semantic prediction from the visible image, and $\mathbf{M}$ be the ground-truth semantic label. 
The visible-failure map is defined as
\begin{gather}
\mathbf{E}_{v}(x)
=
\mathbbm{1}
\left[
\hat{\mathbf{M}}_{v}(x)
\neq
\mathbf{M}(x)
\right],
\end{gather}
where $x$ denotes a pixel. 
We divide visible luminance into $B$ bins. 
For bin $b$, the pixel-level empirical failure is computed as
\begin{gather}
F_b^{\mathrm{pix}}
=
\frac{
\sum_{x\in\Omega_b}\mathbf{E}_{v}(x)
}{
|\Omega_b|+\epsilon
},
\qquad
\Omega_b=\{x\mid \mathbf{Y}_{v}(x)\in \mathcal{B}_b\}.
\end{gather}
To reduce the bias from dominant semantic categories, we also compute a class-normalized failure score:
\begin{gather}
F_b^{\mathrm{cls}}
=
\frac{1}{|\mathcal{C}_b|}
\sum_{c\in\mathcal{C}_b}
\frac{
\sum_{x\in\Omega_b^c}\mathbf{E}_{v}(x)
}{
|\Omega_b^c|+\epsilon
},
\qquad
\Omega_b^c=\{x\mid \mathbf{Y}_{v}(x)\in\mathcal{B}_b,\ \mathbf{M}(x)=c\},
\end{gather}
where $\mathcal{C}_b$ is the set of semantic classes appearing in bin $b$. 
Both empirical failure curves and candidate exposure estimates are normalized to $[0,1]$ before computing MSE.

For each bin, we compare the empirical failure score with the mean analytic prior
\begin{gather}
U_b
=
\frac{1}{|\Omega_b|+\epsilon}
\sum_{x\in\Omega_b}
(2\mathbf{Y}_{v}(x)-1)^2.
\end{gather}
We report Pearson correlation, Spearman correlation, and MSE. 
Given two bin-wise sequences $\mathbf{a}=\{a_b\}_{b=1}^{B}$ and $\mathbf{f}=\{f_b\}_{b=1}^{B}$, Pearson correlation is
\begin{gather}
\rho_{P}
=
\frac{
\sum_{b=1}^{B}(a_b-\bar{a})(f_b-\bar{f})
}{
\sqrt{\sum_{b=1}^{B}(a_b-\bar{a})^2}
\sqrt{\sum_{b=1}^{B}(f_b-\bar{f})^2}
}.
\end{gather}
Spearman correlation is Pearson correlation computed on ranks:
\begin{gather}
\rho_{S}
=
\rho_{P}
\left(
\operatorname{rank}(\mathbf{a}),
\operatorname{rank}(\mathbf{f})
\right).
\end{gather}
The MSE is computed as
\begin{gather}
\operatorname{MSE}
=
\frac{1}{B}
\sum_{b=1}^{B}
(a_b-f_b)^2.
\end{gather}

\cref{tab:app_prior_calibration} and \cref{fig:app_uv_empirical_failure} analyze the relation between exposure estimates and empirical visible failure. 
The two empirical statistics measure different aspects. 
Pixel failure is dominated by frequent pixels and large regions, so it mainly reflects where the visible image actually produces dense prediction errors. 
Class-normalized mFailure gives each semantic class a more balanced contribution within each luminance bin, making it more sensitive to rare classes and category composition. 
Therefore, a prior or posterior can fit one statistic better than the other.

On MSRS, the analytic exposure prior $U_v=(2\mathbf{Y}_{v}-1)^2$ agrees well with class-normalized mFailure. 
It achieves strong class-level Pearson and Spearman correlations and the lowest class-level MSE among analytic estimates. 
This supports the basic physical assumption that both darkness and saturation increase visible unreliability. 
However, the learned posterior $\mathbf{q}_{c}$ is not optimized to reproduce this empirical curve. 
It is a rendering weight calibrated by cross-modal observations. 
As a result, $\mathbf{q}_{c}$ improves pixel-level fitting, especially in MSE, but can be weaker than the raw prior in class-normalized MSE or rank correlation. 
This indicates that the posterior does not simply preserve the exposure prior; it suppresses updates in some exposure-failure regions when infrared structure or local observations do not support context recovery.

The difference is more evident on FMB. 
The class-normalized mFailure on FMB is affected by dataset composition and semantic-category distribution, and the segmentation model used for this post-hoc analysis is not used to train PCDFusion. 
In this setting, darkness-only estimation can correlate better with class-normalized mFailure, while $\mathbf{q}_{c}$ is worse on class-balanced MSE. 
This does not contradict the role of the posterior. 
The context posterior is designed to decide where context rendering should be applied, not to be a class-balanced semantic failure predictor. 
Consistently, $\mathbf{q}_{c}$ achieves the best pixel-level Pearson, Spearman, and MSE on FMB, showing that it better matches dense visible-failure regions relevant to image rendering. 
Overall, the analytic prior provides an interpretable exposure-failure trend, while the posterior further calibrates this trend according to local cross-modal support. 
The calibration is selective rather than uniformly stronger, which is consistent with the goal of avoiding unnecessary context updates in reliable visible regions.

\begin{figure*}[t]
    \centering
    \includegraphics[width=\textwidth]{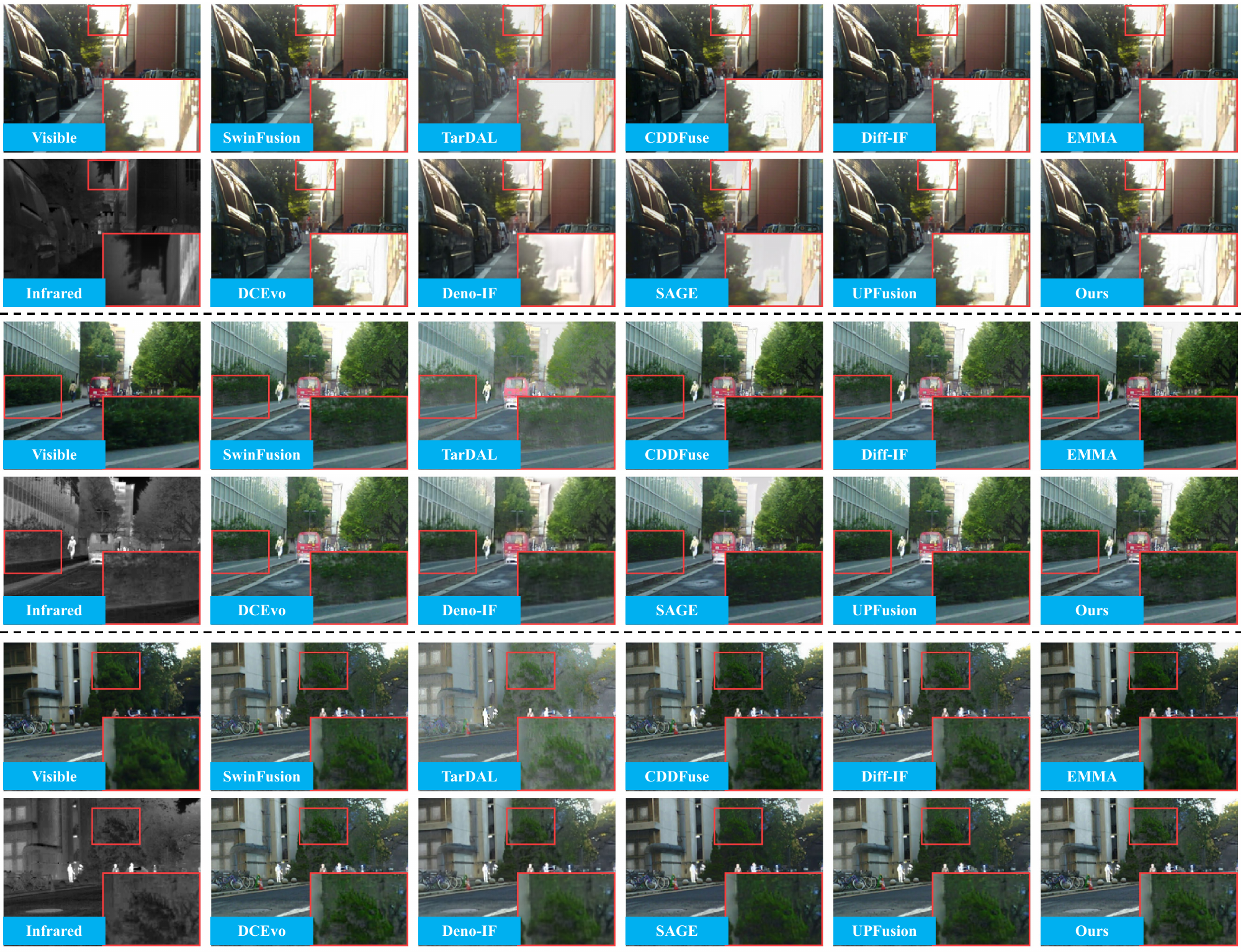}
    \caption{
    Failure cases and limitations. 
    }
    \label{fig:app_failure_cases}
\end{figure*}

\section{Failure Cases and Limitations}
\label{app:failure_cases}

\Cref{fig:app_failure_cases} provides a closer look at the remaining limitations of PCDFusion. 
The first scene shows a typical advantage of our posterior rendering design. 
The visible building region is severely over-exposed, while the infrared image still contains coarse structural cues. 
Conservative methods such as EMMA and UPFusion preserve the visible appearance more strictly, but they also keep the saturated building region largely unresolved. 
In contrast, PCDFusion renders more infrared-supported structure in the failed visible area. 
This result is consistent with the goal of the proposal and context stages, which keep positive thermal responses available and use them to revise exposure-degraded regions.

The second and third scenes reveal the opposite side of this design. 
In these cases, the visible image still contains useful weak structures, while the infrared observation is less reliable for the local appearance. 
Some methods, especially EMMA and UPFusion, behave more conservatively and better preserve the original visible structures with less thermal transfer. 
PCDFusion suppresses obvious infrared contamination more effectively than TarDAL, Diff-IF, and DCEvo, but it still introduces mild thermal interference in the low-luminance regions. 
The effect is visible around weak branches, dark background regions, and low-contrast structures, where infrared responses can be partially rendered even when they do not correspond to reliable visible appearance.

This limitation is related to the role of the context posterior action weight. 
PCDFusion is designed to recover visible failure, so its posterior tends to remain responsive in dark or low-contrast regions. 
As shown in \cref{fig:plot_prior}, the context posterior is relatively high in low-luminance ranges. 
This behavior helps recover genuinely degraded visible regions, but it can also make the model less conservative when the infrared observation is unreliable, weakly aligned, or dominated by smooth thermal background. 
Therefore, the current model still has a trade-off between recovering missing structures and preserving weak but valid visible appearance.

Future work should improve this trade-off in two directions. 
First, the context posterior can incorporate stronger visible-structure preservation, so that low-luminance updates are reduced when visible edges or thin structures remain consistent. 
Second, infrared reliability estimation should explicitly detect cases where thermal responses are likely invalid, such as glass reflection, smooth thermal background, and weakly structured low-light regions. 
A more adaptive posterior calibration may allow the model to keep its advantage in high-exposure recovery while further reducing residual thermal interference in reliable visible regions.

\newpage
\section*{NeurIPS Paper Checklist}

\begin{enumerate}

\item {\bf Claims}
    \item[] Question: Do the main claims made in the abstract and introduction accurately reflect the paper's contributions and scope?
    \item[] Answer: \answerYes{} % Replace by \answerYes{}, \answerNo{}, or \answerNA{}.
    \item[] Justification: The abstract and introduction state the main formulation, the PCDFusion architecture, the absence of semantic labels for fusion training, and the claimed improvements on image-level and downstream tasks. These claims are supported by the method description, quantitative comparisons, ablation studies, and limitations discussion.

\item {\bf Limitations}
    \item[] Question: Does the paper discuss the limitations of the work performed by the authors?
    \item[] Answer: \answerYes{} % Replace by \answerYes{}, \answerNo{}, or \answerNA{}.
    \item[] Justification: The main paper includes a limitations paragraph in the discussion, and the supplemental material provides a separate failure-case analysis. The discussion notes remaining issues in low-luminance and infrared-unreliable regions and relates them to posterior calibration.

\item {\bf Theory assumptions and proofs}
    \item[] Question: For each theoretical result, does the paper provide the full set of assumptions and a complete (and correct) proof?
    \item[] Answer: \answerNA{} % Replace by \answerYes{}, \answerNo{}, or \answerNA{}.
    \item[] Justification: The paper does not claim new theoretical theorems or formal guarantees. It introduces a Bayesian posterior rendering formulation and uses it as a modeling principle for the proposed architecture.

    \item {\bf Experimental result reproducibility}
    \item[] Question: Does the paper fully disclose all the information needed to reproduce the main experimental results of the paper to the extent that it affects the main claims and/or conclusions of the paper (regardless of whether the code and data are provided or not)?
    \item[] Answer: \answerYes{} % Replace by \answerYes{}, \answerNo{}, or \answerNA{}.
    \item[] Justification: The paper reports datasets, splits, training settings, loss functions, model architecture, downstream evaluation protocol, baseline usage, metric implementation, and RGB reconstruction details in the main paper and supplemental material. Code is planned for public release upon acceptance.

\item {\bf Open access to data and code}
    \item[] Question: Does the paper provide open access to the data and code, with sufficient instructions to faithfully reproduce the main experimental results, as described in supplemental material?
    \item[] Answer: \answerNo{} % Replace by \answerYes{}, \answerNo{}, or \answerNA{}.
    \item[] Justification: The experiments use public datasets and public metric code, and the paper states that the implementation will be released upon acceptance. However, the submission does not currently provide an open code repository with executable reproduction scripts.

\item {\bf Experimental setting/details}
    \item[] Question: Does the paper specify all the training and test details (e.g., data splits, hyperparameters, how they were chosen, type of optimizer) necessary to understand the results?
    \item[] Answer: \answerYes{} % Replace by \answerYes{}, \answerNo{}, or \answerNA{}.
    \item[] Justification: The main paper and supplemental material specify the MSRS, FMB, and M3FD splits, training schedule, optimizer, learning rate, batch size, crop size, downstream fine-tuning details, baseline checkpoint usage, and metric computation.


\item {\bf Experiment statistical significance}
    \item[] Question: Does the paper report error bars suitably and correctly defined or other appropriate information about the statistical significance of the experiments?
    \item[] Answer: \answerNo{} % Replace by \answerYes{}, \answerNo{}, or \answerNA{}.
    \item[] Justification: The paper reports deterministic dataset-level averages over standard test splits but does not include repeated runs, confidence intervals, or statistical significance tests. This choice follows common IVIF evaluation practice, but the absence of error bars is a limitation.


\item {\bf Experiments compute resources}
    \item[] Question: For each experiment, does the paper provide sufficient information on the computer resources (type of compute workers, memory, time of execution) needed to reproduce the experiments?
    \item[] Answer: \answerYes{} % Replace by \answerYes{}, \answerNo{}, or \answerNA{}.
    \item[] Justification: The supplemental material reports the GPU type, CPU configuration, training GPU memory, per-epoch training time, and single-image inference memory. It also reports model parameters, GFLOPs, inference time, and FPS under a fixed input size.
    
\item {\bf Code of ethics}
    \item[] Question: Does the research conducted in the paper conform, in every respect, with the NeurIPS Code of Ethics \url{https://neurips.cc/public/EthicsGuidelines}?
    \item[] Answer: \answerYes{} % Replace by \answerYes{}, \answerNo{}, or \answerNA{}.
    \item[] Justification: The work uses public benchmark datasets and does not involve private data collection, human-subject experiments, or manipulative user-facing systems. To the best of our knowledge, the research conforms to the NeurIPS Code of Ethics.

\item {\bf Broader impacts}
    \item[] Question: Does the paper discuss both potential positive societal impacts and negative societal impacts of the work performed?
    \item[] Answer: \answerNo{} % Replace by \answerYes{}, \answerNo{}, or \answerNA{}.
    \item[] Justification:  The paper focuses on the technical formulation and evaluation of IVIF and does not include a dedicated broader-impact discussion. Potential benefits include improved perception under challenging illumination, while possible risks include use in surveillance or safety-critical systems without sufficient validation.
    
\item {\bf Safeguards}
    \item[] Question: Does the paper describe safeguards that have been put in place for responsible release of data or models that have a high risk for misuse (e.g., pre-trained language models, image generators, or scraped datasets)?
    \item[] Answer: \answerNA{} % Replace by \answerYes{}, \answerNo{}, or \answerNA{}.
    \item[] Justification: The paper does not release a high-risk pretrained generative model, language model, or scraped dataset. It uses existing public IVIF benchmarks and proposes an image fusion model.

\item {\bf Licenses for existing assets}
    \item[] Question: Are the creators or original owners of assets (e.g., code, data, models), used in the paper, properly credited and are the license and terms of use explicitly mentioned and properly respected?
    \item[] Answer: \answerYes{} % Replace by \answerYes{}, \answerNo{}, or \answerNA{}.
    \item[] Justification: The supplemental material lists the public repositories or sources of all datasets, baseline implementations, metric code, and downstream pretrained checkpoints used in the experiments. It also reports the visible repository-level license or terms when available, and marks assets as not specified when no explicit license is shown on the public repository page.

\item {\bf New assets}
    \item[] Question: Are new assets introduced in the paper well documented and is the documentation provided alongside the assets?
    \item[] Answer: \answerNA{} % Replace by \answerYes{}, \answerNo{}, or \answerNA{}.
    \item[] Justification: The paper does not introduce a new dataset or benchmark. The implementation code is planned for release upon acceptance, but no new asset package is provided with the current anonymous submission.


\item {\bf Crowdsourcing and research with human subjects}
    \item[] Question: For crowdsourcing experiments and research with human subjects, does the paper include the full text of instructions given to participants and screenshots, if applicable, as well as details about compensation (if any)? 
    \item[] Answer: \answerNA{} % Replace by \answerYes{}, \answerNo{}, or \answerNA{}.
    \item[] Justification: The paper does not involve crowdsourcing, user studies, or research with human participants.

\item {\bf Institutional review board (IRB) approvals or equivalent for research with human subjects}
    \item[] Question: Does the paper describe potential risks incurred by study participants, whether such risks were disclosed to the subjects, and whether Institutional Review Board (IRB) approvals (or an equivalent approval/review based on the requirements of your country or institution) were obtained?
    \item[] Answer: \answerNA{} % Replace by \answerYes{}, \answerNo{}, or \answerNA{}.
    \item[] Justification: The paper does not involve crowdsourcing or human-subject research, so IRB approval is not applicable.

\item {\bf Declaration of LLM usage}
    \item[] Question: Does the paper describe the usage of LLMs if it is an important, original, or non-standard component of the core methods in this research? Note that if the LLM is used only for writing, editing, or formatting purposes and does \emph{not} impact the core methodology, scientific rigor, or originality of the research, declaration is not required.
    %this research? 
    \item[] Answer: \answerNA{} % Replace by \answerYes{}, \answerNo{}, or \answerNA{}.
    \item[] Justification: LLMs are not used as an important, original, or non-standard component of the proposed method, experiments, or evaluation.

\end{enumerate}

\end{document}